\documentclass[1p,review]{elsarticle}

\usepackage{hyperref}
\usepackage{lineno}

\usepackage{graphicx}
\usepackage{subfigure}
\usepackage{amsmath,amssymb}
\usepackage{multirow}
\usepackage{algorithm}
\usepackage[noend]{algorithmic}

\usepackage[utf8]{inputenc}

\journal{Pattern Recognition}

\begin{document}
	
	\begin{frontmatter}
		
		\title{Two-stage generative adversarial networks for document image binarization with color noise and background removal}
		
		\author[KISTEurope,TUKaiserslautern]{Sungho Suh}
		\ead[url]{https://github.com/opensuh/DocumentBinarization/}
		\author[KISTEurope,KHU]{Jihun Kim}
		\author[TUKaiserslautern,DFKI]{Paul Lukowicz}
		\author[KISTEurope]{Yong Oh Lee\corref{mycorrespondingauthor}}
		\cortext[mycorrespondingauthor]{Corresponding author}
		\ead{yongoh.lee@kist-europe.de}

		\address[KISTEurope]{Smart Convergence Group, Korea Institute of Science and Technology Europe Forschungsgesellschaft mbH, 66123 Saarbrücken, Germany}
		\address[TUKaiserslautern]{Department of Computer Science, TU Kaiserslautern, 67663 Kaiserslautern, Germany}
		\address[KHU]{Department of Computer Science and Engineering, Kyung Hee University, 17104 Yongin-si, South Korea}
		\address[DFKI]{German Research Center for Artificial Intelligence (DFKI), 67663 Kaiserslautern, Germany}

		\begin{abstract}
			Document image enhancement and binarization methods are often used to improve the accuracy and efficiency of document image analysis tasks such as text recognition. Traditional non-machine-learning methods are constructed on low-level features in an unsupervised manner but have difficulty with binarization on documents with severely degraded backgrounds. Convolutional neural network (CNN)––based methods focus only on grayscale images and on local textual features. In this paper, we propose a two-stage color document image enhancement and binarization method using generative adversarial neural networks. In the first stage, four color-independent adversarial networks are trained to extract color foreground information from an input image for document image enhancement. In the second stage, two independent adversarial networks with global and local features are trained for image binarization of documents of variable size. For the adversarial neural networks, we formulate loss functions between a discriminator and generators having an encoder--decoder structure. Experimental results show that the proposed method achieves better performance than many classical and state-of-the-art algorithms over the Document Image Binarization Contest (DIBCO) datasets, the LRDE Document Binarization Dataset (LRDE DBD), and our shipping label image dataset. We plan to release the shipping label dataset as well as our implementation code at \texttt{github.com/opensuh/DocumentBinarization/}.			
		\end{abstract}
		
		\begin{keyword}
			Document image binarization \sep Generative adversarial networks \sep Optical character recognition \sep Color document image enhancement
		\end{keyword}
		
	\end{frontmatter}
		
	\section{Introduction}
	\label{introduction}
	
	Document image enhancement is one of the most important pre-processing steps for the tasks involved in analyzing an imaged document. The main aim of document image enhancement is to extract the foreground text from the degraded background in the document image, a process called image binarization. The better binarization methods will improve the performance of subsequent document analysis tasks \cite{michalak2018region} such as layout analysis \cite{antonacopoulos2011historical}, text line and word segmentation \cite{stamatopoulos2013icdar}, and optical character recognition (OCR) \cite{smith2007overview}. The challenge is that document images normally suffer from various types of degradation and interference, such as page stains, uneven illumination, contrast variation, ink bleed, paper yellowing, background color, bleed-through, and environmental deterioration \cite{kligler2018document, sulaiman2019degraded}. 
	
	Over the past few years, many document image enhancement methods have been proposed in the literature, such as those that improve image quality by removing degradation effects and artifacts present in an image to restore its original appearance \cite{moghaddam2009variational, hedjam2013historical}. However, traditional unsupervised document enhancement methods rarely handle multiple degradations. Most focus on a single specific problem, such as bleed-through \cite{yagoubi2015new, sun2016blind}.
	
	Recently, deep-learning-based image binarization methods have provided significant improvements over traditional image processing methods, not simply solving a specific problem but eliminating multiple degradations \cite{otsu1979threshold, niblack1986introduction, sauvola2000adaptive}. A selectional auto-encoder model for document image binarization was proposed by Calvo-Zaragoza and Gallego \cite{calvo2019selectional}. To generate binarized output the same size as the input image, a fully convolutional neural network (FCN) was used, generating binary results as a special case of semantic segmentation \cite{peng2013exploiting, long2015fully, tensmeyer2017document}. To improve the performance of the binarization for a degraded document, several approaches have been proposed, including a hierarchical deep supervised network \cite{vo2018binarization}, an iterative supervised network \cite{he2019deepotsu}, and a conditional generative-adversarial-network-based binarization method \cite{zhao2019document}, which outperformed traditional methods and other deep-learning-based methods. 
	
	\begin{figure}[!t]
		\centering
		\subfigure[]{
			\includegraphics[width=\columnwidth]{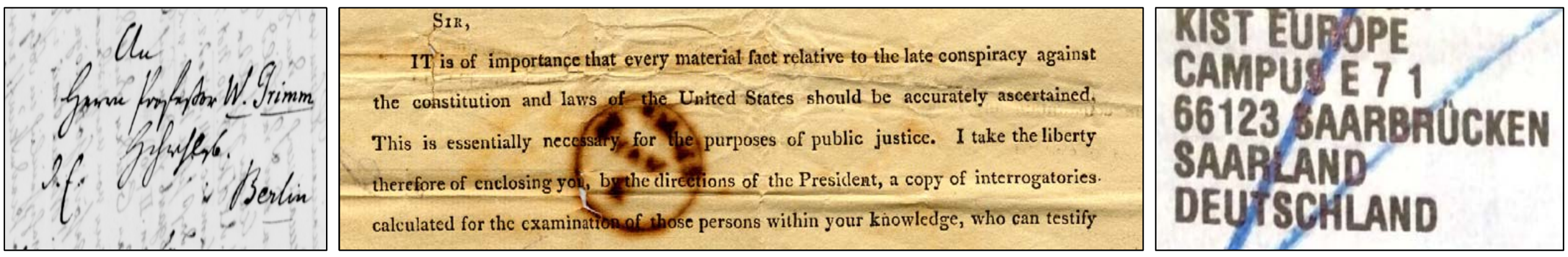}
		}
		\hfil
		\subfigure[]{
			\includegraphics[width=\columnwidth]{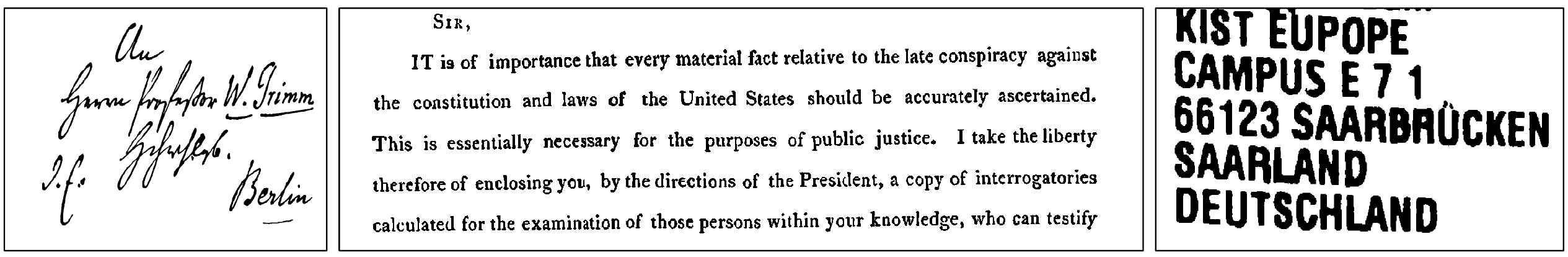}
		}
		\hfil
		\subfigure[]{
			\includegraphics[width=\columnwidth]{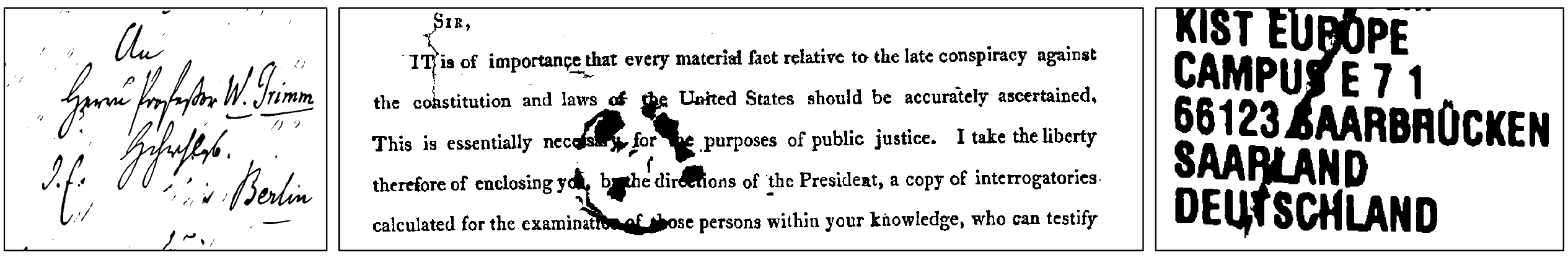}
		}
		\caption{Examples of the degraded document and their challenges in image binarization. (a) some of the example images from DIBCO dataset and shipping label image dataset, (b) the corresponding ground truth images, (c) the corresponding binarization results by Vo's method \cite{vo2018binarization}.}
		\label{fig:documentimageexample}
	\end{figure}
	
	However, all of these methods focus only on grayscale document images because the scanned historical documents that most of them are designed to handle are contaminated document images in black and white. Although these methods remove colorless degradation factors well, they are weaker in extracting the target from colorful backgrounds and in removing colored contaminants. Furthermore, most of the methods are based solely on context information in a particular neighborhood region, using small local image patches instead of the whole image as the input to the deep learning network. For regions of densely placed text, local features in local image patches are advantageous for extracting text from backgrounds, whereas the spatial contextual information can sometimes be missed for regions with a largely blank background. These two major problems are exemplified in Figure \ref{fig:documentimageexample}.
	
	Taking these problems into consideration, we propose a framework of generative adversarial networks for color document image enhancement and binarization. The proposed method consists of two stages, the first of which trains multiple color channel adversarial neural networks to extract color foreground information from small local image patches by removing background information for document image enhancement, and the second of which learns the local and global binary results with multi-scale adversarial neural networks for document image binarization. In the first stage, four color-independent networks are trained with red, green, blue, and gray channel images, respectively. In the second stage, the local binary transformation network is trained with the merged enhanced images from the first stage, and the global binary transformation network is trained with the full original input image. 
	
	
	To evaluate the proposed method on degraded document images, we used three types of document image binarization datasets: the Document Image Binarization Contest (DIBCO) datasets \cite{gatos2009icdar, pratikakis2010h, 6065249, pratikakis2012icfhr, pratikakis2013icdar, ntirogiannis2014icfhr2014, pratikakis2016icfhr2016}, the LRDE Document Binarization Dataset (LRDE DBD) \cite{lazzara2011scribo, lazzara2014efficient}, and a shipping label image dataset \cite{suh2019robust, suh2020fusion}.
	
	The contributions of this paper can be summarized as follows. (1) Unlike previous methods, the proposed method focuses on the multi-color degradation problem with its design of four color-independent networks. (2) By combining local and global binary transformation networks, the method can balance the fine extraction of strokes and the suppression of background misclassification. (3) The proposed method is evaluated on multiple datasets and is found to achieve better performance than state-of-the-art models. 
	
	The rest of the paper is organized as follows. Section \ref{sec:relatedwork} summarizes related work. Section \ref{sec:proposedmethod} provides the details of the proposed method. Section \ref{sec:experimentalresults} presents quantitative and qualitative experimental results on various datasets. Finally, Section \ref{sec:conclusion} concludes the paper.
	
	\section{Related Work}\label{sec:relatedwork}
	
	Studies of document image binarization have been reported in the literature over the past two decades. The goal of binarization is to convert each pixel of an input document image into either text or background with two levels of gray. Generally, document image binarization methods can be categorized into two types: traditional non-machine-learning algorithms and semantic segmentation methods based on deep learning. The most well-known classical global binarization method is Otsu's method \cite{otsu1979threshold}, which computes a threshold that minimizes the within-class variance and maximizes the between-class variance of two pre-assumed classes. It selects the global threshold based on a histogram, and with this simplified calculation, the operation is fast. However, the classical global binarization methods are sensitive and unstable with noise, non-uniform backgrounds, and degraded documents. To solve these problems, local adaptive binarization methods, such as those of Niblack \cite{niblack1986introduction}, Sauvola and Pietik{\"a}inen \cite{sauvola2000adaptive}, and Wolf and Jolion \cite{wolf2004extraction}, have been proposed. These methods compute the pixel-wise local threshold based on local statistical information. 
	
	Following these methods, several binarization methods have been proposed. Gatos et al. \cite{gatos2006adaptive} used a low-pass Wiener filter to estimate the background and foreground regions in a method that employs several post-processing steps to remove noise in the background and improve the quality of foreground regions. Su et al. \cite{su2012robust} introduced an adaptive contrast map combining the local image contrast and the local image gradient; the local threshold is estimated based on the values on the detected edges in a local region. Pai et al. \cite{pai2010adaptive} proposed an adaptive window-size selection method based on the image characteristic; the threshold is determined in each window region. Howe \cite{howe2013document} formulated the binarization task as a global energy loss, the parameters of which can be tuned automatically. Jia et al. \cite{jia2018degraded} employed the structural symmetry of strokes to compute the local threshold. However, these local adaptive binarization methods require a number of empirical parameters and are still not satisfactory for use with highly degraded and poor-quality document images.
	
	In recent years, convolutional neural networks (CNNs) have achieved significant improvements in a variety of tasks in machine learning and computer vision applications \cite{krizhevsky2012imagenet, zeiler2014visualizing, lecun2015deep}. Since document image binarization is regarded as an image segmentation task, object segmentation methods in deep learning are widely applied. In \cite{tensmeyer2017document}, a fully convolutional network (FCN) is applied for the document image binarization task at multiple image scales. A convolutional deep auto-encoder--decoder architecture model is used for binarization in \cite{peng2017using, calvo2019selectional}. A hierarchical deep supervised network has been proposed to predict the full foreground map through the results of three multi-scale networks \cite{vo2018binarization}; this method achieved state-of-the-art performance on several benchmark datasets. He and Schomaker \cite{he2019deepotsu} proposed an iterative deep learning framework and achieved performance similar to that of Vo et al.'s method \cite{vo2018binarization}.
	
	Recently, generative adversarial networks (GANs) \cite{goodfellow2014generative} have emerged as a class of generative models approximating real data distributions, achieving promising performance in the generation of real-world images \cite{konwer2018staff, suh2019generative, suh2020cegan}. The original purpose of GAN \cite{goodfellow2014generative} was to train generative models to capture real data distributions and to generate an output image from random noise. A generator network competes against a discriminator network that distinguishes between generated and real images. Unlike the original GAN, CGAN \cite{mirza2014conditional} trains the generator not only to fool the discriminator but also to condition on additional inputs, such as class labels, partial data, or input images.
	Isola et al. \cite{isola2017image} proposed Pix2Pix GAN for the general purpose of image-to-image translation using CGAN. The generator of the Pix2Pix GAN model is trained via adversarial loss, which encourages the generation of plausible images in the target domain and minimizes the measured L1 loss between the generated image and the expected output image. The discriminator is provided with both the source image and a target image and determines whether the target image is a plausible transformation of the source image.
	
	\begin{equation}
		\label{eq:pix2pix}
		\begin{split}
			\min_{G} \max_{D} \mathop{\mathbb{E}_{x,y}}[\log D(y,x)] + \mathop{\mathbb{E}_{x}}[\log (1-D(G(x), x))] + \lambda \mathop{\mathbb{E}_{x,y}}[\Arrowvert y - G(x) \Arrowvert_1],
		\end{split}
	\end{equation}
	where $x$ is an input image sampled from the input data distribution $\mathop{\mathbb{P}_x}$, $y$ is the ground truth image corresponding to the input image, $\lambda$ is a hyperparameter that increases the effect of regularization on a model, and the generator $G$ generates an image $G(x)$ from an input image. Equation (\ref{eq:pix2pix}) shows that minimization of L1 distance, rather than L2 distance, between the generated image and the ground truth image encourages less blurring and ambiguity in the generation process \cite{isola2017image}. Pix2Pix GAN was demonstrated on a range of image-to-image translation tasks such as conversions of maps to satellite photographs, black-and-white photographs to color, and product sketches to product photographs. 
	
	Konwer et al. \cite{konwer2018staff} used Pix2Pix GAN to remove the staff line for optical music recognition. Bhunia et al. \cite{bhunia2019improving} proposed a texture augmentation network to augment the training datasets and handle image binarization by using a CGAN structure. Zhao et al. \cite{zhao2019document} proposed a cascaded generator structure based on Pix2Pix GAN to combine global and local information, and De et al. \cite{de2020document} introduced a dual discriminator GAN containing two discriminator networks to combine global and local information as well. They achieved results comparable to those of state-of-the-art methods on DIBCO datasets. 
	
	\section{Proposed Method}\label{sec:proposedmethod}
	
	\begin{figure}[!t]
		\centering
		\includegraphics[width=\columnwidth]{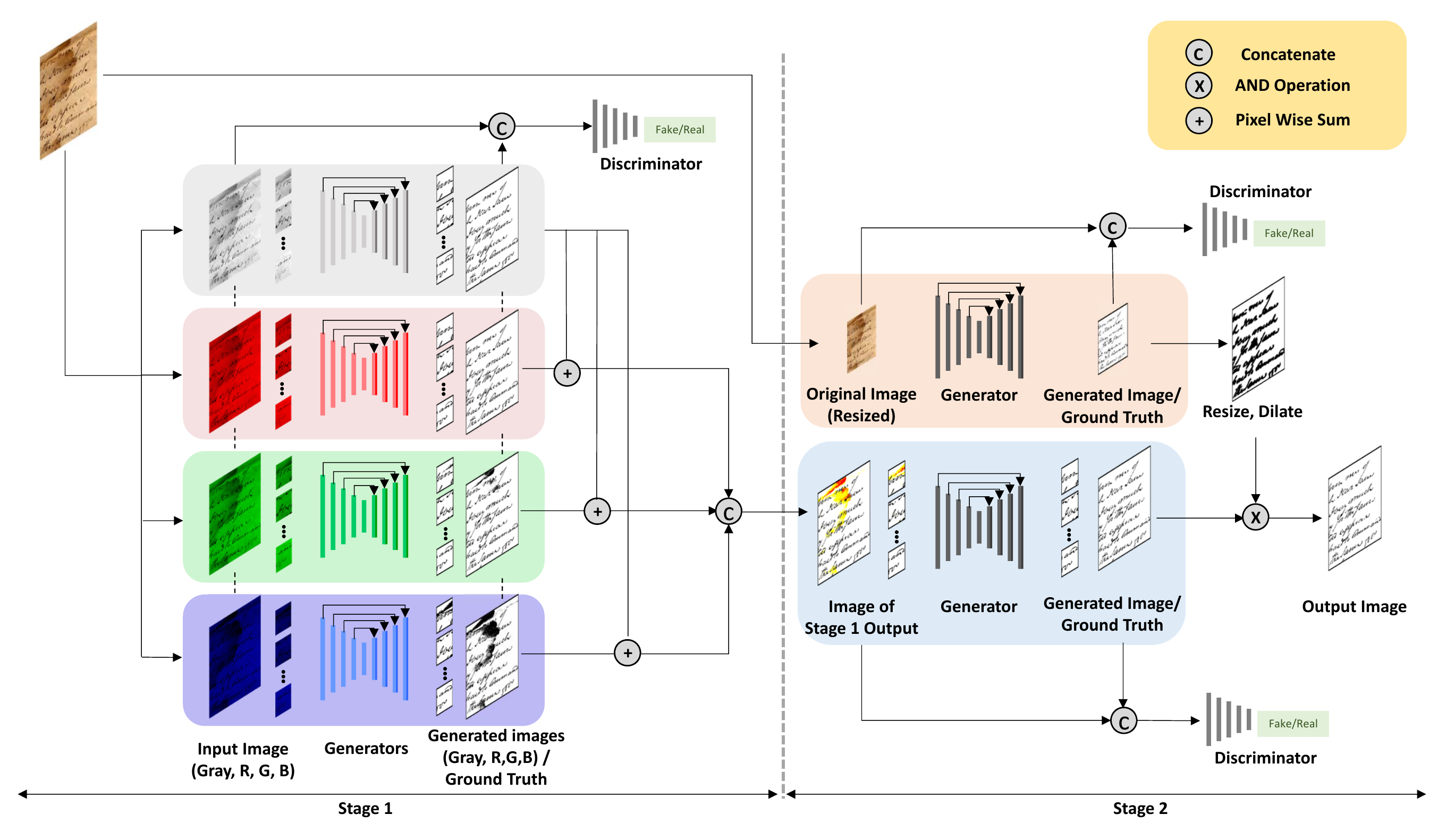}
		\caption{The structure of the proposed model for document image enhancement (Stage 1) and document image binarization (Stage 2). }
		\label{fig:fig_overview}
	\end{figure}
	
	\textbf{Loss functions of the proposed GAN}
	Although GAN and CGAN are capable of generating fake images close to the original input images or target images, the GAN training procedure has instability of loss function convergence \cite{goodfellow2016nips}. To solve this stability problem, we apply the Wasserstein GAN with gradient penalty (WGAN-GP) \cite{gulrajani2017improved}, which uses the Wasserstein-K distance as the loss function, to the objective function to guide the training process. Furthermore, unlike the general image-to-image transfer task \cite{isola2017image}, our task, document image enhancement and binarization, is to primarily classify every individual pixel into two classes, text and background. Thus, we decided to use the binary cross-entropy (BCE) loss rather than L1 loss as used in previous approaches \cite{isola2017image, zhao2019document}. Bartusiak et al. \cite{bartusiak2019splicing} showed experimentally that BCE loss is indeed a better choice than L1 loss for binary classification. The objective loss functions of conditional WGAN-GP with the BCE loss are defined as follows:
	
	\begin{equation}
		\label{eq:dloss}
		\begin{split}
			\mathop{\mathbb{L}_D}(x,y;\theta_D) = - \mathop{\mathbb{E}_{x,y}}[D(y,x)] + \mathop{\mathbb{E}_{x}}[D(G(x), x)] + \alpha \mathop{\mathbb{E}_{x, \hat{y}\sim P_{\hat{y}}}}[(\Arrowvert \nabla_{\hat{y}} D(\hat{y}, x) \Arrowvert_2 -1 )^2],
		\end{split}
	\end{equation}
	\begin{equation}
		\label{eq:gloss}
		\begin{split}
			\mathop{\mathbb{L}_G}&(x,y;\theta_G) = \mathop{\mathbb{E}_{x}}[D(G(x),x)] + \lambda \mathop{\mathbb{L}_{BCE}}(G(x), y),\\
			&\text{where}~\mathop{\mathbb{L}_{BCE}}(p,q)= \mathop{\mathbb{E}_{p,q}} [q\log p + (1-q)\log (1-p)],
		\end{split}
	\end{equation}
	where $\alpha$ is the penalty coefficient; $P_{\hat{y}}$ is the uniform sampling along straight lines between pairs of points from the ground truth distribution $P_y$ and the generated data distribution; $\lambda$ controls the relative importance of different loss terms; and $\theta_D$ and $\theta_G$ are parameters of the discriminator and the generator, respectively.  Whereas the discriminator $D$ is trained to minimize $\mathop{\mathbb{L}_D}$ for distinguishing between the ground truth and the generated image, the generator $G$ is trained to minimize $\mathop{\mathbb{L}_G}$.
	
	\textbf{Network architecture of the proposed GAN} 
	The GAN architecture has two neural networks, and we selected a generator and a discriminator for image binarization performance. As the encoder in the generator, U-Net \cite{ronneberger2015u} is employed. U-Net is a network structure that introduces skip concatenation between the encoder and the decoder layers; it provides good performance in image segmentation. The encoder includes downsampling to extract the context information, and the decoder is an upsampling process that combines the upsampled features and the low-dimensionality features from the downsampling layer to improve the performance of the network. In binarization studies, U-Net is widely adopted \cite{isola2017image, he2019deepotsu, zhao2019document, de2020document}. To extract important features effectively in the proposed adversarial neural networks, we adopted EfficientNet \cite{tan2019efficientnet} as the encoder in the generator; it has achieved much better accuracy and efficiency for image classification than other networks. As the discriminator, the discriminator network in PatchGAN \cite{li2016precomputed, isola2017image, zhu2017unpaired, ledig2017photo, zhao2019document, de2020document} is employed with the modification. PatchGAN has been used frequently in recent work due to its good generalization properties. We employ a network with an architecture similar to that of the discriminator network from the discriminator, based on a Markov random field model, of Pix2Pix GAN \cite{isola2017image}. 

	The overall network architecture of the proposed method is shown in Figure \ref{fig:fig_overview}. The first stage of the proposed method, shown on the left side of the figure, consists of four color-independent generators and a discriminator to distinguish between the ground truth and the generated image. Four color-independent networks are trained with red, green, blue, and gray channel images, respectively, and produce an enhanced document image by removing background color information. Each channel image and the corresponding ground truth or generated image are concatenated and fed into the discriminator. In the second stage, shown on the right side of the figure, the local binary transformation network is trained with the merged enhanced images resulting from the first stage, and the global binary transformation network is trained with the full original input image for document image binarization.
	
	\subsection{Document Image Enhancement using Color-Independent Adversarial Networks}
	\label{subsec:enhancement}
	
	Our goal is to enhance the degraded color document and extract the text regions from the image. Owing to the diversity and complexity of the degraded color document images, it is difficult to extract the text regions in the variety of conditions of degradation and printing by training only one generator or by training with a grayscale image. Instead of generating a binary image directly from a three-channel color input image or a grayscale input image, we first train four color-independent generators, which focus on extracting color information and removing the color background. Four color-independent networks are trained with split three-channel images and a converted grayscale image. 
	
	Before training the proposed GAN framework in the first stage, a pre-processing step for ground truth images of three color input images is required. If a given ground truth image were to be used directly to train four different generators, the corresponding networks would not be able to extract the proper color information and could not be trained to classify the text regions and the background regions from the corresponding color channel owing to the lack of information. To generate ground truth images for three different color images, we combine a split input channel image and a given ground truth image via the logical \textsc{and} operator. After combining the ground truth image and the split channel image, we generate a binary image via application of the global threshold. Unlike the three split color images, the given ground truth image is used directly for training the grayscale generator with the converted grayscale image. 
	
	After the pre-processing step for ground truth images, the four generators and one discriminator are trained to extract colored text regions from the background regions with the input images and their corresponding ground truth images. During training, we divide each document image into small patches of size $256 \times 256$ without resizing. The four generators are trained to fool the discriminator and minimize the distance between the generated image and the ground truth image in each color channel by minimizing the objective loss function $\mathop{\mathbb{L}_G}$ in (\ref{eq:gloss}), and the discriminator is trained to distinguish between the generated image and the ground truth image by minimizing the objective loss function $\mathop{\mathbb{L}_D}$ in (\ref{eq:dloss}). 
	
	The four generated images $G_{r}(x_r)$, $G_g(x_g)$, $G_b(x_b)$, and $G_{gray}(x_{gray})$ are predicted for each image patch via the four proposed generators. To integrate the information in the four images predicted by the four color channel generators, we apply pixel-wise addition between $G_{gray}(x_{gray})$ and the three generated color images $G_{r}(x_r)$, $G_g(x_g)$, and $G_b(x_b)$, respectively, and merge the three channel images into a color image. The training details for the proposed method are summarized in Algorithm \ref{Step1Algo}.

	\begin{algorithm}[ht]
		\caption{Training procedure for document image enhancement using adversarial networks. We use default values of $\omega = 0.5$, $\alpha = 10$, $\lambda=50$}\label{Step1Algo}
		\begin{algorithmic}[1]
			\REQUIRE Batch size $m$, Adam hyperparameters $\eta$, hyperparameter for $\lambda$.
			\STATE \textbf{Initialize:} $\theta_{G_r}$, $\theta_{G_g}$, $\theta_{G_b}$, $\theta_{G_{gray}}$ from pre-trained source networks.
			\FOR{number of training iterations}
			\STATE Sample $\{x^{(i)}\}^m_{i=1}$ a batch from the input image patches and corresponding ground truth $\{y^{(i)}\}^m_{i=1}$.		\FOR{$k=\{r, g, b, gray\}$}
			\IF{$k$ is not gray}
			\STATE $y_k \gets x_k \bigcap y$
			\STATE Binarize $y_k$ with threshold value $t$.
			\ELSE
			\STATE $y_k \gets y$
			\ENDIF
			\STATE Update discriminator $D$ by descending the gradient of (\ref{eq:dloss}):
			\STATE $\theta_D \gets \theta_D - \eta_D \nabla_{\theta_D} \mathop{\mathbb{L}_D}(x_k,y_k;\theta_D)$
			\STATE Update generator $G_k$ by descending the gradient of (\ref{eq:gloss}):
			\STATE $\theta_{G_k} \gets \theta_{G_k} - \eta_G \nabla_{\theta_{G_k}} \mathop{\mathbb{L}_G}(x_k,y_k;\theta_{G_k})$
			\ENDFOR
			\ENDFOR
			\FOR{$k=\{r, g, b\}$}
			\STATE $\hat{y}_k \gets \omega G_k(x_k) + (1-\omega) G_{gray}(x_{gray})$
			\ENDFOR
			\STATE $\hat{y} \gets [\hat{y}_{r}, \hat{y}_{g}, \hat{y}_{b}]$
		\end{algorithmic}
	\end{algorithm}
	
	\subsection{Document Image Binarization using Multi-scale Feature Fusion}
	\label{subsec:binarization}
	As shown on the right side of Figure \ref{fig:fig_overview}, the second stage combines the global and local results of the document binarization. The binarization in the first stage is performed mainly using local prediction with the small patches. However, for a document image having a large background region portion, local prediction can sometimes misclassify parts of the background region as foreground. To address this problem, we perform the global binarization with the resized original input image and the local binarization with the image result from the first stage. The reason for using the resized original input image is that the image result from the first stage can have loss of spatial contextual information of the document image since the first stage is only performed using local prediction. The input image for the generator and the binary image, which is the corresponding ground truth image or the generated image, are concatenated and fed into the discriminator. Whereas the input image for the generators is an 8-bit image in the first stage, in the second stage it is a 24-bit three-channel image. Furthermore, the input image for the local prediction is the small patches of the output image from the first stage, whereas the input image for the global prediction is the resized original input image. Thus, the second stage for multi-scale feature fusion requires two discriminators, whereas the first stage trains only one shared discriminator with four independent generators. 
	
	For the local and global binarizations, the network architecture of the generator is the same as that in the first stage except for the input image channel. The degradation document image datasets, such as DIBCO and the shipping label dataset, contain images of various sizes and height-to-width ratios. In the local binarization process, regions of different colors can be removed from text components, and text-only regions can be extracted by using the image result from the document image enhancement stage. For the global binarization, we add padding to the whole image or directly resize the original image to the desired size $(r,r)$. The strategies of global binarization are as follows according to the ratio of the input image height $h$ and width $w$:
	
	\begin{itemize}[]
		\item \textbf{If} $ \frac{\max(w, h)}{\min(w, h)} < 4.0$, resize the whole image to $(r,r)$.
		\item \textbf{If} $4.0 \leq \frac{\max(w, h)}{\min(w, h)} < 6.0$, put the input image at the center of $(\max(w, h),\max(w, h))$, add padding with the median pixel value of the input image set to $(\max(w, h),\max(w, h))$, and resize the image of $(\max(w,h), \max(w,h))$ to $(r,r)$.
		\item \textbf{If} $\frac{\max(w, h)}{\min(w, h)} \geq 6.0$ or $(h < r$ and $w < r)$, do not perform global binarization.
	\end{itemize}
	
	Morphological dilation is applied to the global prediction map so that the global map can cover all of the text regions because the resizing process can remove textual components. Lastly, the global binarization image is incorporated into the local binarization map via the logical \textsc{and} operator to remove any misclassified noise in the background. 
		
	\section{Experimental Results}\label{sec:experimentalresults}
	\subsection{Datasets and Evaluation Metrics}
	\label{subsec:dataset}
	To evaluate the proposed method on degraded document images, three types of document image binarization datasets were used: 
	
	\begin{itemize}[]
		\item DIBCO: Seven competition datasets---DIBCO 2009 \cite{gatos2009icdar}, H-DIBCO 2010 \cite{pratikakis2010h}, DIBCO 2011 \cite{6065249}, H-DIBCO 2012 \cite{pratikakis2012icfhr}, DIBCO 2013 \cite{pratikakis2013icdar}, H-DIBCO 2014 \cite{ntirogiannis2014icfhr2014}, and H-DIBCO 2016 \cite{pratikakis2016icfhr2016}---were selected for the experiments. A total of 86 degraded document images were used. 
		\item LRDE DBD \cite{lazzara2011scribo, lazzara2014efficient}: LRDE DBD is a dataset of French magazine images consisting of 125 color document images. For evaluation, we conducted a five-fold cross-validation procedure for LRDE DBD. 
		\item A shipping label image dataset \cite{suh2019robust, suh2020fusion}: The shipping label image dataset consists of 1082 images of different types and from various countries, acquired using smartphones. For evaluation, we conducted a five-fold cross-validation procedure for these as well.
	\end{itemize}
	
	To compare the performance of the proposed method with state-of-the-art algorithms, we constructed training and testing sets from the DIBCO datasets similar to those described in \cite{vo2018binarization, he2019deepotsu}. A total of 34 images from DIBCO 2009, H-DIBCO 2010, and H-DIBCO 2012 were selected for training. The training set also contained a total of 109 images from the Persian Heritage Image Binarization Dataset (PHIBD) \cite{nafchi2013efficient}, the Synchromedia Multispectral Ancient Document Images (SMADI) dataset \cite{hedjam2013historical}, and the Bickley diary dataset \cite{deng2010binarizationshop}. A total of 52 images from DIBCO 2011, DIBCO 2013, H-DIBCO 2014, and H-DIBCO 2016 were used for testing. 
	
	For quantitative evaluation and comparison with the state-of-the-art algorithms, we adopted four evaluation metrics, which are used in the image binarization competition DIBCO: F-measure (FM), pseudo-F-measure (p-FM), peak signal-to-noise ratio (PSNR), and the distance reciprocal distortion (DRD) metric. Additionally, to evaluate the degree to which the proposed method improves OCR performance on degraded document images, we adopted the Levenshtein distance \cite{levenshtein1966binary} expressed in percent. We measured the Levenshtein distance only for images from the shipping label image dataset, for which the ground truth is given.
	
	\begin{itemize}
		\item \textbf{F-measure (FM)}: 
		\begin{equation}
			\label{eq:FM}
			\begin{split}
				&FM = \frac{2\times Recall \times Precision}{Recall + Precision},
			\end{split}
		\end{equation}
		where $Recall=\frac{TP}{TP+FN}$, $Precision=\frac{TP}{TP+FP}$, and $TP$, $FP$, and $FN$ denote the true positive, false positive, and false negative values, respectively.
		\item \textbf{Pseudo-F-measure (p-FM)}:
		\begin{equation}
			\label{eq:pFM}
			\begin{split}
				&p-FM = \frac{2\times pRecall \times Precision}{pRecall + Precision},
			\end{split}
		\end{equation}
		where $pRecall$ is the percentage of the skeletonized ground truth image \cite{pratikakis2010h}.
		\item \textbf{Peak signal-to-noise ratio (PSNR)}:
		\begin{equation}
			\label{eq:psnr}
			\begin{split}
				&PSNR = 10\log\left(\frac{C^2}{MSE}\right),
			\end{split}
		\end{equation}
		where $MSE=\frac{\sum_{x=1}{M}\sum_{y=1}^{N}(L(x,y)-L'(x,y))^2}{MN}$, and $C$ denotes the difference between text and background. $PSNR$ is a measure of the similarity between two images.
		\item \textbf{Distance reciprocal distortion (DRD) metric}:
		\begin{equation}
			\label{eq:DRD}
			\begin{split}
				&DRD= \frac{\sum_k DRD_k}{NUBN},
			\end{split}
		\end{equation}
		where $DRD_k$ is the distortion of the $k$th flipped pixel, and $NUBN$ is the number of non-uniform $8\times 8$ blocks in the ground truth image \cite{ntirogiannis2014icfhr2014}.
		\item \textbf{Levenshtein distance expressed in percent (Lev)}:
		\begin{equation}
			\label{eq:Lev}
			\begin{split}
				&Lev(s_1,s_2)= \left(1-\frac{d(s_1,s_2)}{\max(\arrowvert s_1 \arrowvert, \arrowvert s_2 \arrowvert)}\right)\times 100,
			\end{split}
		\end{equation}
		where $d(s_1,s_2)$ denotes the Levenshtein distance between two strings $s_1$, $s_2$ (of length $\arrowvert s_1 \arrowvert$ and $\arrowvert s_2 \arrowvert$, respectively).
	\end{itemize}
	
	\subsection{Implementation Details}
	The experiments were all implemented using Python scripts in the PyTorch framework. Training procedures were conducted in the Linux system with NVIDIA Tesla V100 GPUs.
	
	\textbf{Data preparation} To compare just the performance of the algorithms, we used the same training sets and data augmentations for the proposed method and the other state-of-the-art methods. We cut the images in the training datasets into small patches ($256 \times 256$) to train the local prediction networks. Data augmentation was applied to create more training samples. For the DIBCO datasets, we sampled patches with the scale factors \{0.75, 1.25, 1.5\}, resized the patches to $256 \times 256$, and rotated with a rotation angle of 270$^{\circ}$. Overall, around 120,000 training image patches were created from the DIBCO datasets. For LRDE DBD, scale and rotation augmentations were not utilized and only a horizontal flip was applied, but more than 310,000 training patches were created. Lastly, for the shipping label image dataset, we applied only scale augmentation, with the scale factors \{0.75, 1.25, 1.5\}. Approximately 65,000 small patches were used in the training procedure. To train the global prediction network, we determined the desired size $(r,r)$ to be $(512, 512)$ and followed the strategies listed in Section \ref{subsec:binarization}. Data augmentation was also applied for global prediction. We applied the rotation augmentation with rotation angles \{90$^{\circ}$, 180$^{\circ}$, 270$^{\circ}$\}, and horizontal and vertical flip.

	\textbf{Training} Our parameter settings for the first and second stages were the same except for the number of training global epochs, which were 150. The proposed networks were trained for 10 epochs on all datasets. We chose the Adam optimizer with a learning rate of $2\times 10^{-4}$, $\beta_1=0.5$, and $\beta_2=0.999$ for generator and discriminator. We plan to release the shipping label dataset as well as our implementation code at \texttt{github.com/opensuh/DocumentBinarization/}.
	
	\textbf{Pre-training} We initialized the encoder in the generator with weights pre-trained on the ImageNet dataset \cite{deng2009imagenet} to overcome the shortage of training data.
	
	\subsection{Comparison with State-of-the-art Methods}
	The proposed method was evaluated on the DIBCO datasets, LRDE DBD, and the shipping label image dataset. The five evaluation metrics defined in Section \ref{subsec:dataset} were used to evaluate and compare the proposed method with traditional binarization methods, including Otsu's \cite{otsu1979threshold}, Niblack's \cite{niblack1986introduction}, and Sauvola and Pietik{\"a}inen's \cite{sauvola2000adaptive} methods, and with deep-learning-based state-of-the-art methods, including Vo et al.'s \cite{vo2018binarization}, He and Schomaker's \cite{he2019deepotsu}, and Zhao et al.'s \cite{zhao2019document} methods. The traditional methods convert the three-channel input images into grayscale images. Vo et al.'s and He and Schomaker's methods also feed the grayscale images as input images to the neural networks; only Zhao et al.'s method uses the original three-channel input images as the input images. Additionally, Vo et al.'s and Zhao et al.'s methods combine the local and global prediction results, whereas the other traditional methods and He and Schomaker's method only use the global prediction results from the whole input image or the local prediction result from small patch images. 
	
	\subsubsection{Results on DIBCO Datasets}
	
	The quantitative evaluation on the DIBCO datasets was performed on four datasets: DIBCO 2011, DIBCO 2013, H-DIBCO 2014, and H-DIBCO 2016. Because the DIBCO datasets have handwritten texts and do not provide the ground truth for the OCR outputs, we evaluated the proposed method and the state-of-the-art methods by using four evaluation metrics (omitting the Levenshtein distance). 
	
	Table \ref{tab:dibco} shows the quantitative evaluation results on the four DIBCO datasets. On DIBCO 2011 and H-DIBCO 2016 (Table \ref{tab:dibco} parts (a) and (d)), the proposed method performed the best in terms of all four of the measurements. On DIBCO 2013 and H-DIBCO 2014, (Table \ref{tab:dibco} parts (b) and (c)), the proposed method achieved the best performance in terms of FM, PSNR, and DRD but ranked third in terms of p-FM. The FM and DRD of the proposed method are markedly better than those of the state-of-the-art methods. These results imply that the proposed method differentiates textual components from background more effectively than the state-of-the-art methods and is more robust. 
	
	\begin{table}
		\caption{Evaluation of document image binarization on the DIBCO datasets.}
		\label{tab:dibco}
		\centering
		\subfigure[DIBCO 2011]{
			\resizebox{0.45\columnwidth}{!}{
				\begin{tabular}{ccccc}
					\hline
					Methods & FM & p-FM & $PSNR$ & $DRD$\\
					\hline
					Otsu \cite{otsu1979threshold} & 82.10 & 85.96 & 15.72 & 8.95\\
					Niblack \cite{niblack1986introduction} & 70.44 & 73.03 & 12.39 & 24.95 \\
					Sauvola \cite{sauvola2000adaptive} & 82.35 & 88.63 & 15.75 & 7.86 \\
					Vo \cite{vo2018binarization} & 92.58 & 94.67 & 19.16 & 2.38 \\
					He \cite{he2019deepotsu} & 91.92 & 95.82 & 19.49 & 2.37 \\
					Zhao \cite{zhao2019document} & 92.62 & 95.38 & 19.58 & 2.55 \\
					Ours & \textbf{93.57} & \textbf{95.93} & \textbf{20.22} & \textbf{1.99}\\
					\hline
			\end{tabular}}
		}
		\subfigure[DIBCO 2013]{
			\resizebox{0.45\columnwidth}{!}{
				\begin{tabular}{ccccc}
					\hline
					Methods & FM & p-FM & $PSNR$ & $DRD$\\
					\hline
					Otsu \cite{otsu1979threshold} & 80.04 & 83.43 & 16.63 & 10.98\\
					Niblack \cite{niblack1986introduction} & 71.38 & 73.17 & 13.54 & 23.10 \\
					Sauvola \cite{sauvola2000adaptive} & 82.73 & 88.37 & 16.98 & 7.34 \\
					Vo \cite{vo2018binarization} & 93.43 & 95.34 & 20.82 & 2.26 \\
					He \cite{he2019deepotsu} & 93.36 & \textbf{96.70} & 20.88 & 2.15 \\
					Zhao \cite{zhao2019document} & 93.86 & 96.47 & 21.53 & 2.32 \\
					Ours & \textbf{95.01} & 96.49 & \textbf{21.99} & \textbf{1.76}\\
					\hline
			\end{tabular}}
		}
		\hfil
		\subfigure[H-DIBCO 2014]{
			\resizebox{0.45\columnwidth}{!}{
				\begin{tabular}{ccccc}
					\hline
					Methods & FM & p-FM & $PSNR$ & $DRD$\\
					\hline
					Otsu \cite{otsu1979threshold} & 91.62 & 95.69 & 18.72 & 2.65\\
					Niblack \cite{niblack1986introduction} & 86.01 & 88.04 & 16.54 & 8.26 \\
					Sauvola \cite{sauvola2000adaptive} & 83.72 & 87.49 & 17.48 & 5.05 \\
					Vo \cite{vo2018binarization} & 95.97 & 97.42 & 21.49 & 1.09 \\
					He \cite{he2019deepotsu} & 95.95 & \textbf{98.76} & 21.60 & 1.12 \\
					Zhao \cite{zhao2019document} & 96.09 & 98.25 & 21.88 & 1.20 \\
					Ours & \textbf{96.36} & 97.87 & \textbf{21.96} & \textbf{1.07}\\
					\hline
			\end{tabular}}
		}
		\subfigure[H-DIBCO 2016]{
			\resizebox{0.45\columnwidth}{!}{
				\begin{tabular}{ccccc}
					\hline
					Methods & FM & p-FM & $PSNR$ & $DRD$\\
					\hline
					Otsu \cite{otsu1979threshold} & 86.59 & 89.92 & 17.79 & 5.58\\
					Niblack \cite{niblack1986introduction} & 72.57 & 73.51 & 13.26 & 24.65 \\
					Sauvola \cite{sauvola2000adaptive} & 84.27 & 89.10 & 17.15 & 6.09 \\
					Vo \cite{vo2018binarization} & 90.01 & 93.44 & 18.74 & 3.91 \\
					He \cite{he2019deepotsu} & 91.19 & 95.74 & 19.51 & 3.02 \\
					Zhao \cite{zhao2019document} & 89.77 & 94.85 & 18.80 & 3.85 \\
					Ours & \textbf{92.24} & \textbf{95.95} & \textbf{19.93} & \textbf{2.77}\\
					\hline
			\end{tabular}}
		}
	\end{table}

	\begin{figure}[!t]
		\centering
		\subfigure[]{
			\includegraphics[width=0.15\columnwidth]{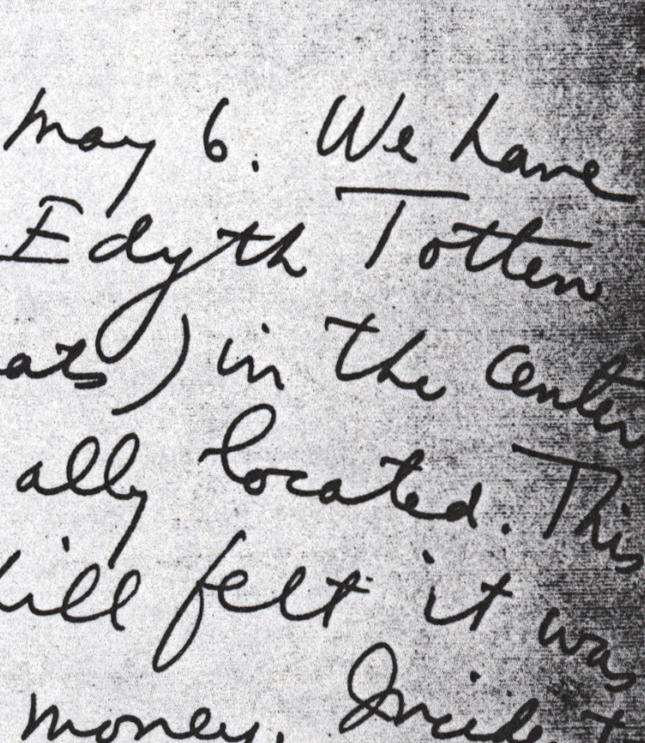}
		}
		\subfigure[]{
			\includegraphics[width=0.15\columnwidth]{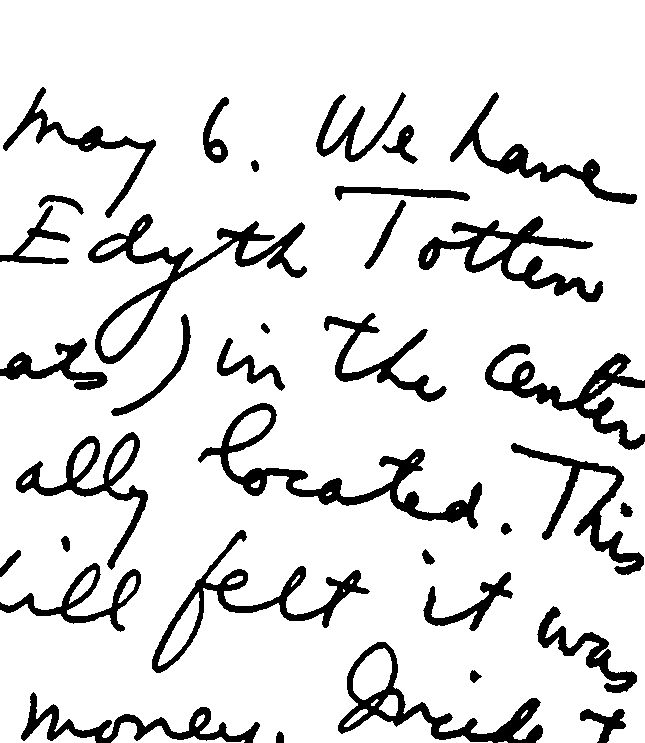}
		}
		\subfigure[]{
			\includegraphics[width=0.15\columnwidth]{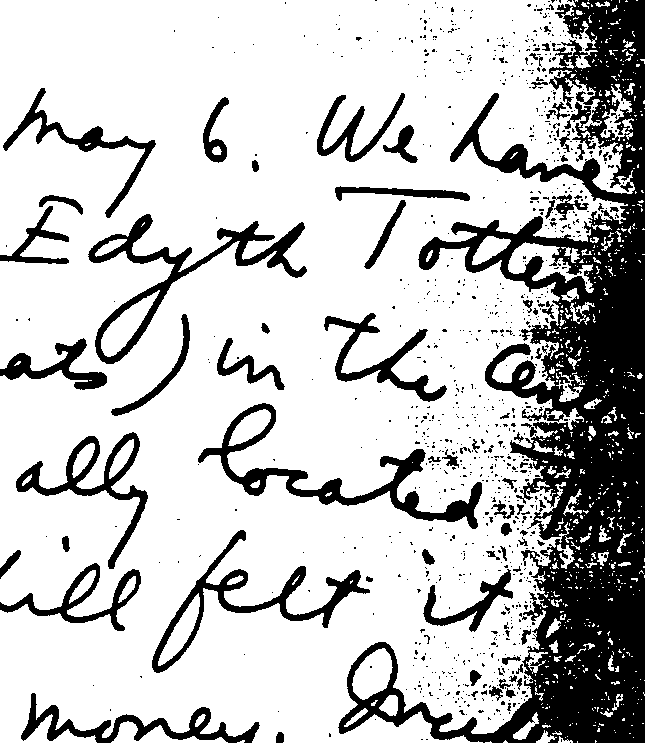}
		}
		\hskip 0.5\columnwidth
		\subfigure[]{
			\includegraphics[width=0.15\columnwidth]{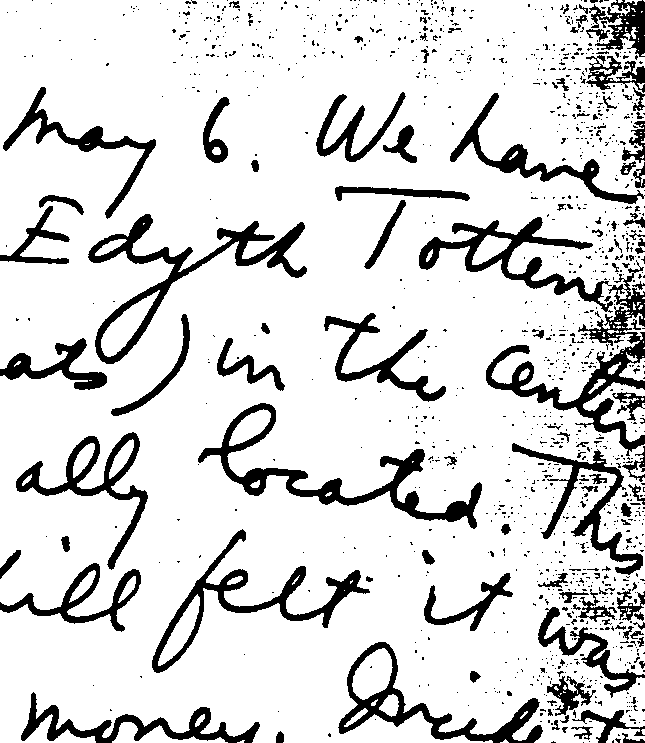}
		}
		\subfigure[]{
			\includegraphics[width=0.15\columnwidth]{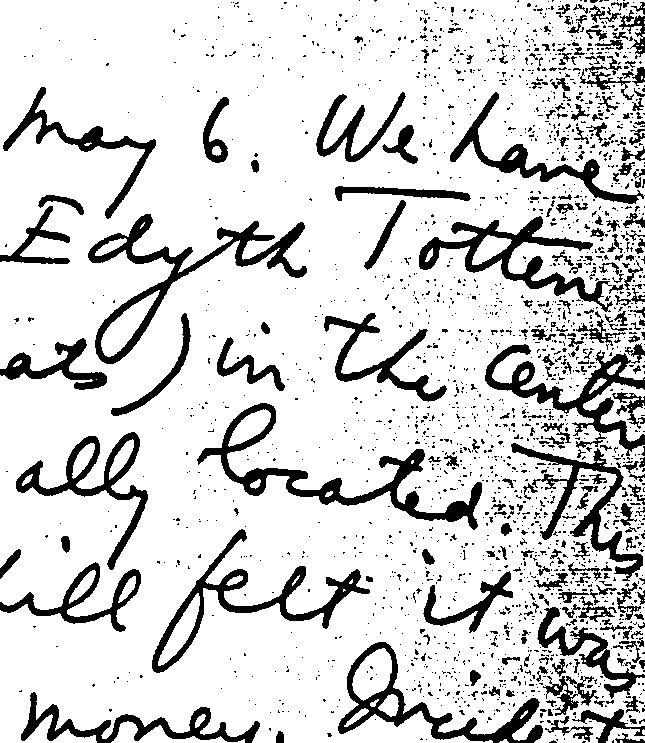}
		}
		\subfigure[]{
			\includegraphics[width=0.15\columnwidth]{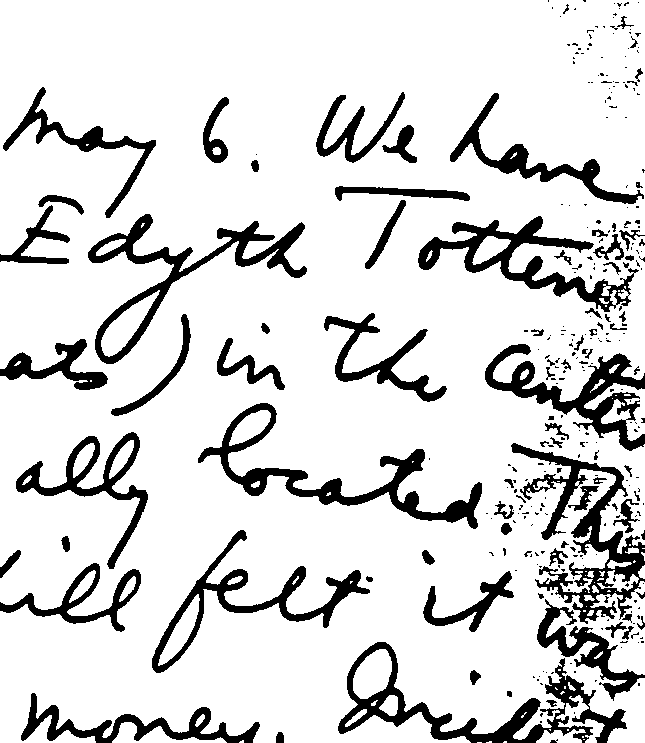}
		}
		\hskip 0.5\columnwidth
		\subfigure[]{
			\includegraphics[width=0.15\columnwidth]{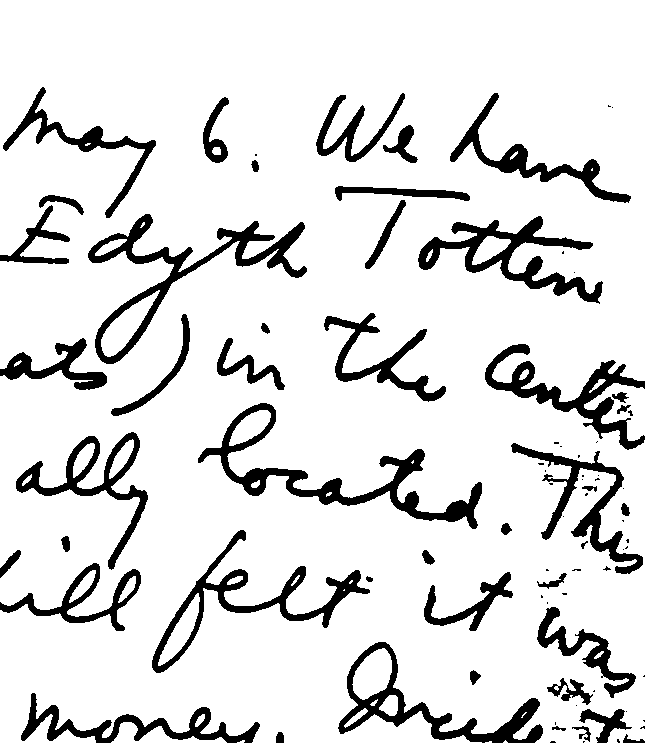}
		}
		\subfigure[]{
			\includegraphics[width=0.15\columnwidth]{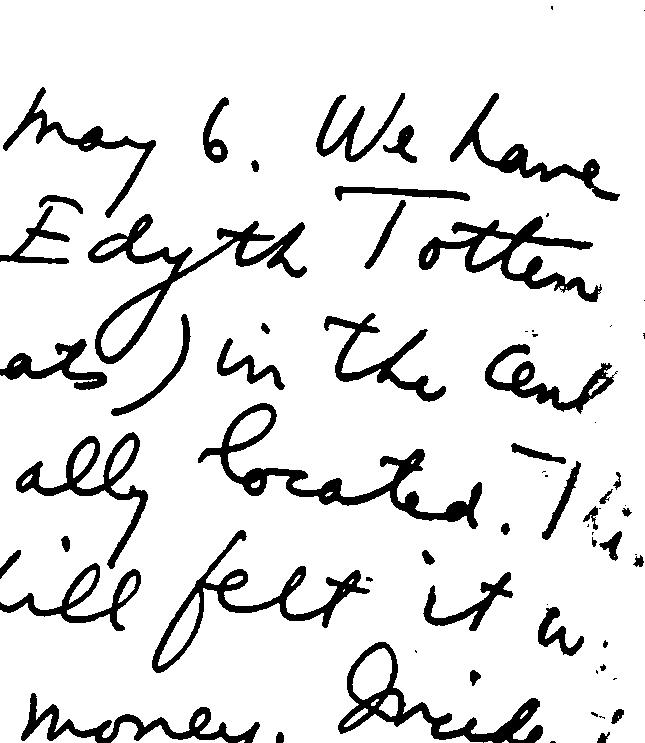}
		}
		\subfigure[]{
			\includegraphics[width=0.15\columnwidth]{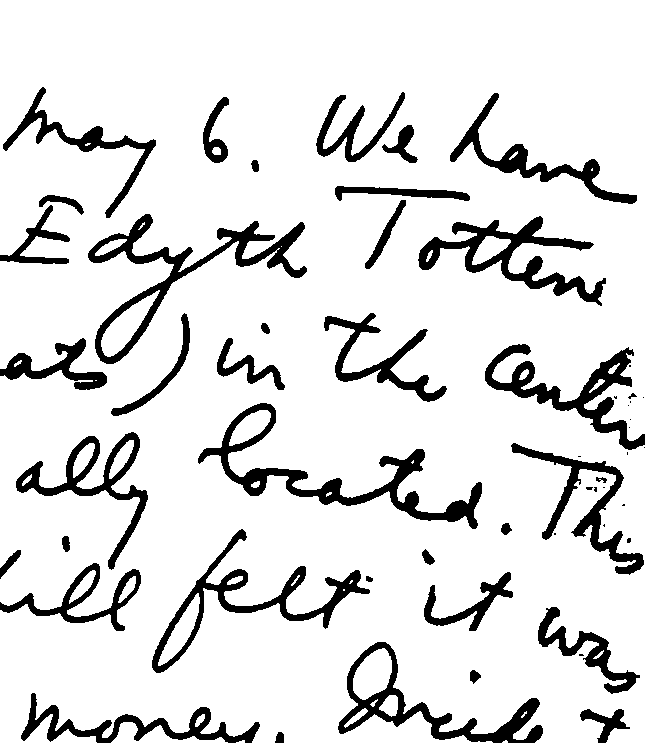}
		}
		\caption{Binarization results of the document image HW1 in DIBCO 2011. (a) original images, (b) the ground truth, (c) Otsu \cite{otsu1979threshold}, (d) Niblack \cite{niblack1986introduction}, (e) Sauvola \cite{sauvola2000adaptive}, (f) Vo \cite{vo2018binarization}, (g) He \cite{he2019deepotsu}, (h) Zhao \cite{zhao2019document}, (i) Ours.}
		\label{fig:DIBCO2011_1}
	\end{figure}
	\begin{figure}
		\centering
		\subfigure[]{
			\includegraphics[width=0.15\columnwidth]{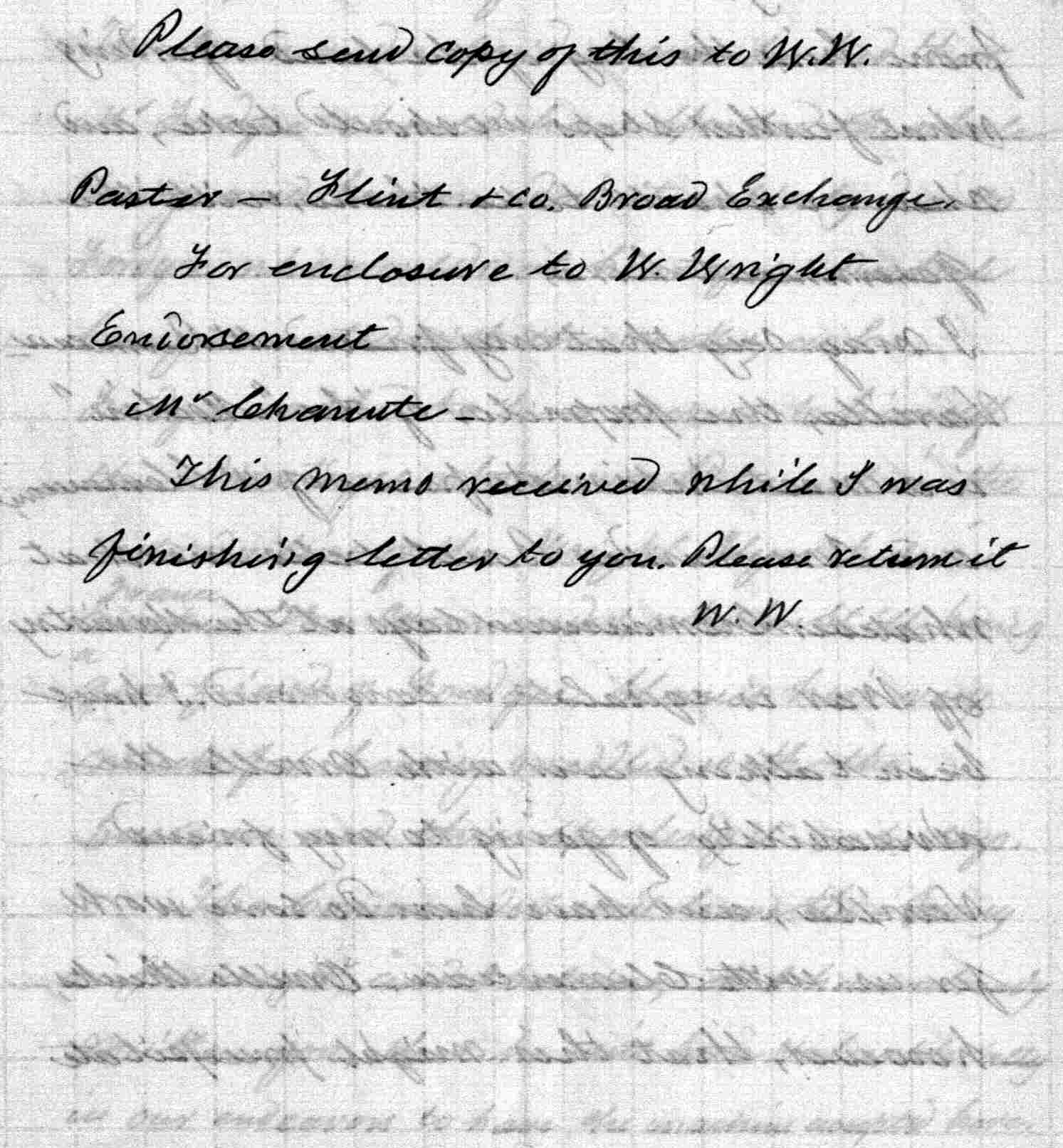}
		}
		\subfigure[]{
			\includegraphics[width=0.15\columnwidth]{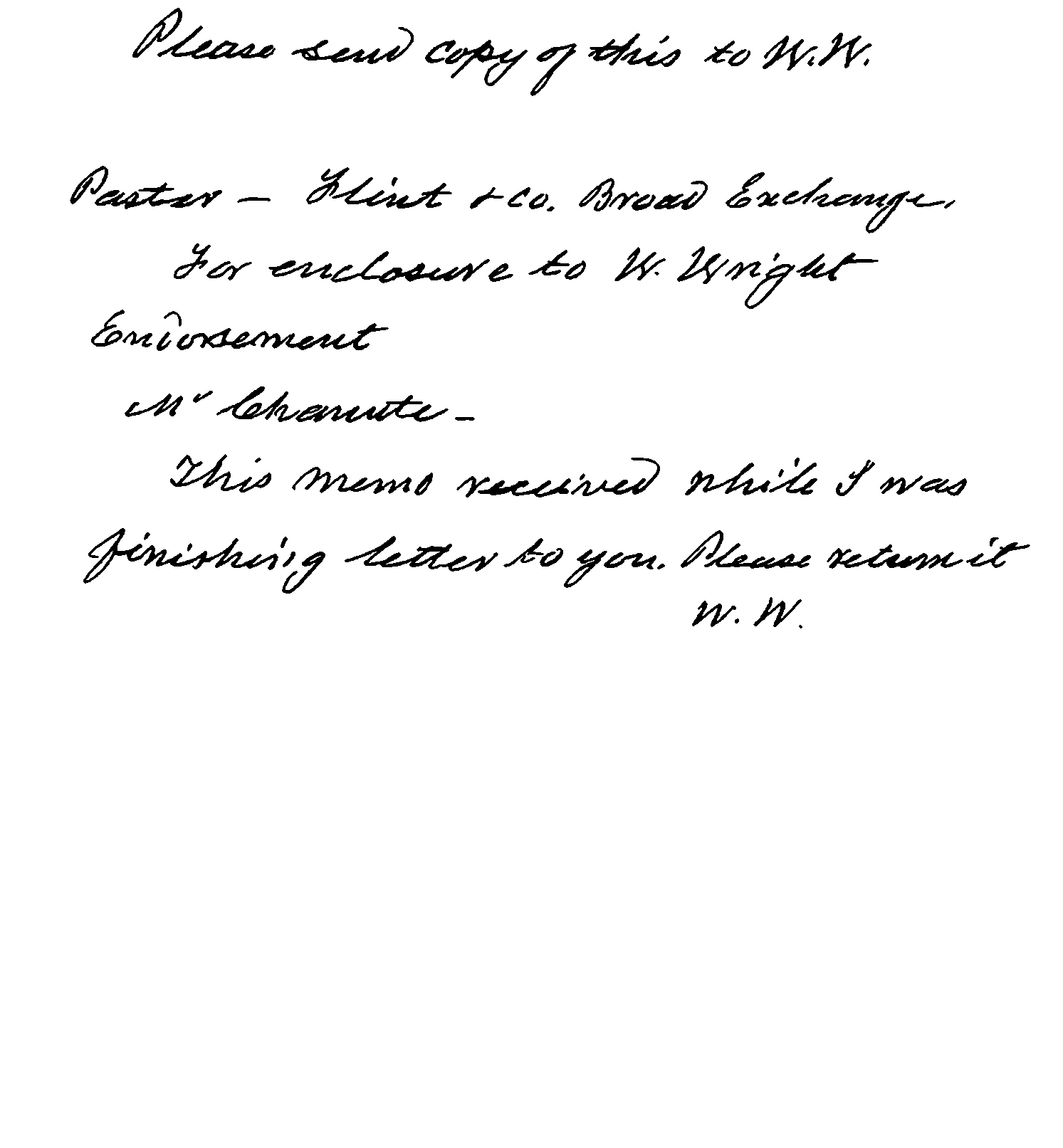}
		}
		\subfigure[]{
			\includegraphics[width=0.15\columnwidth]{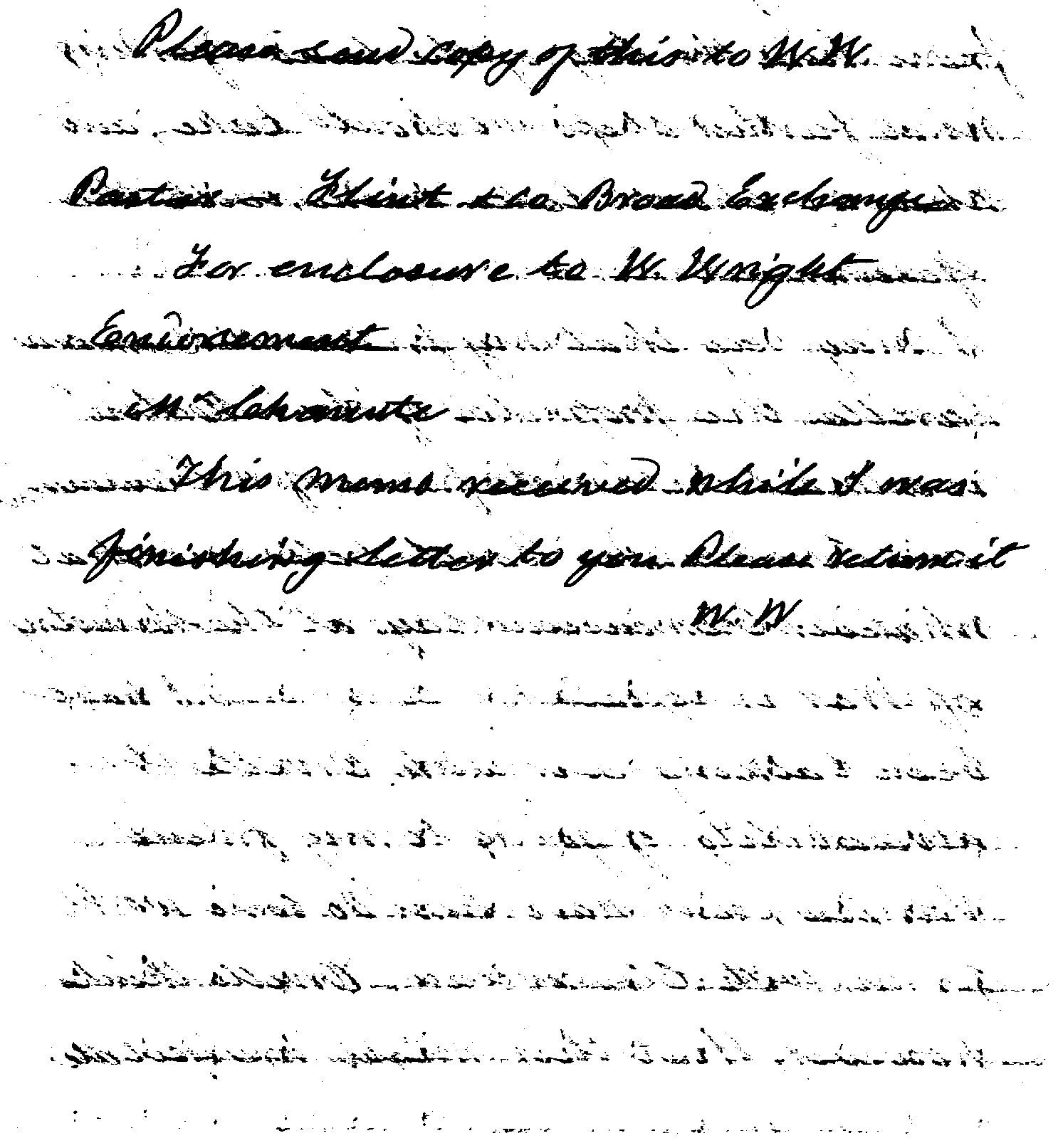}
		}
		\hskip 0.5\columnwidth
		\subfigure[]{
			\includegraphics[width=0.15\columnwidth]{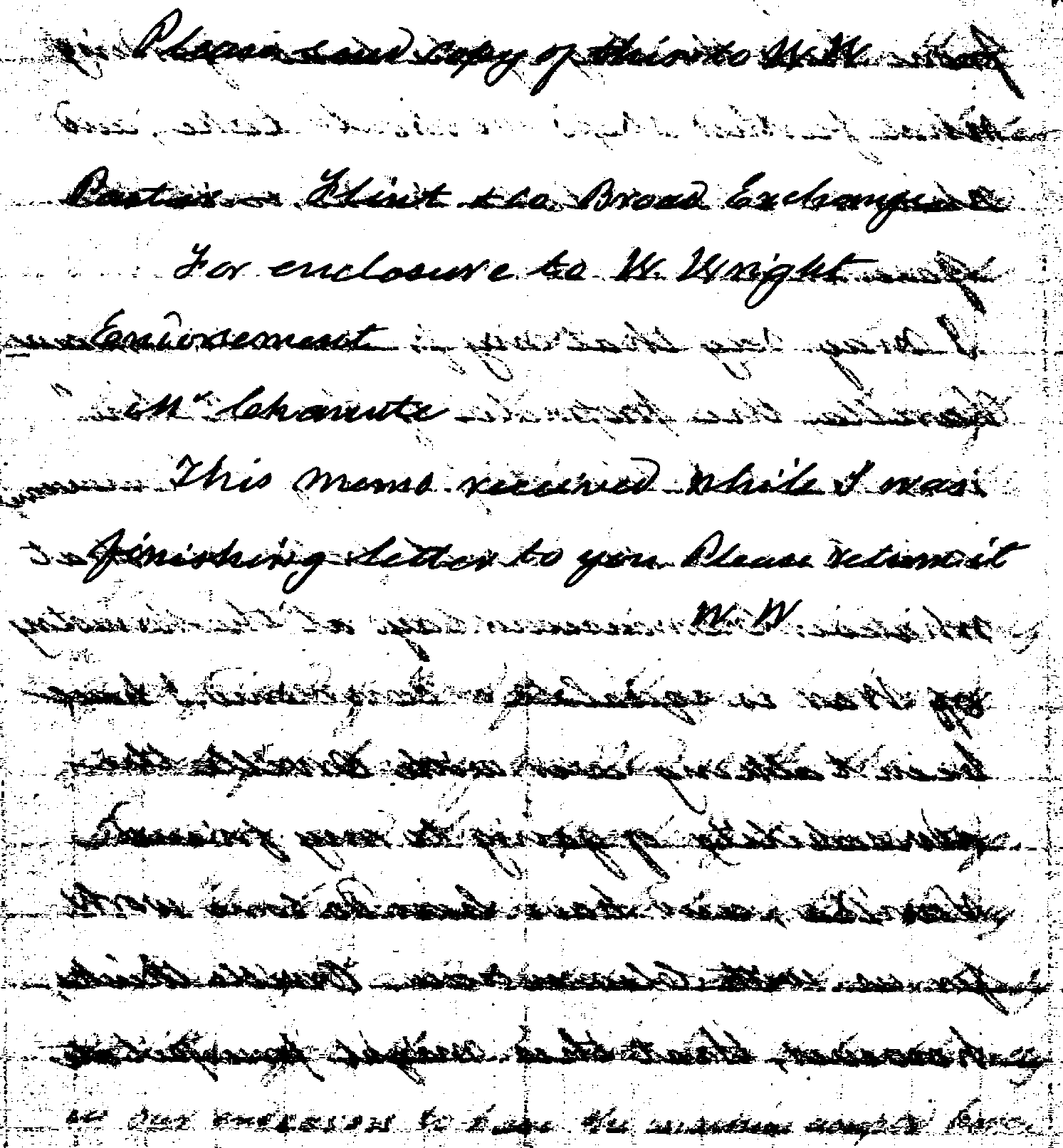}
		}
		\subfigure[]{
			\includegraphics[width=0.15\columnwidth]{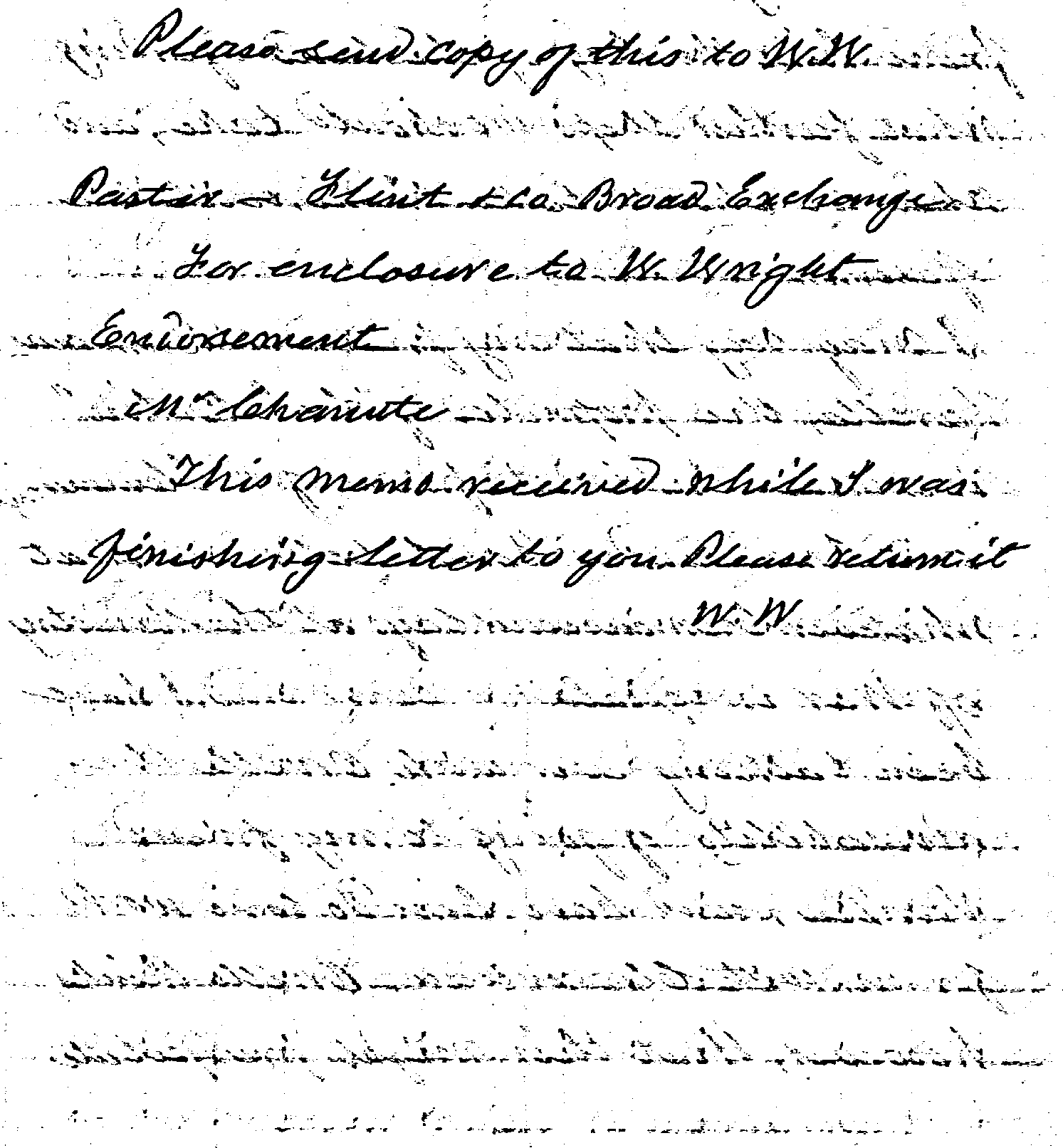}
		}
		\subfigure[]{
			\includegraphics[width=0.15\columnwidth]{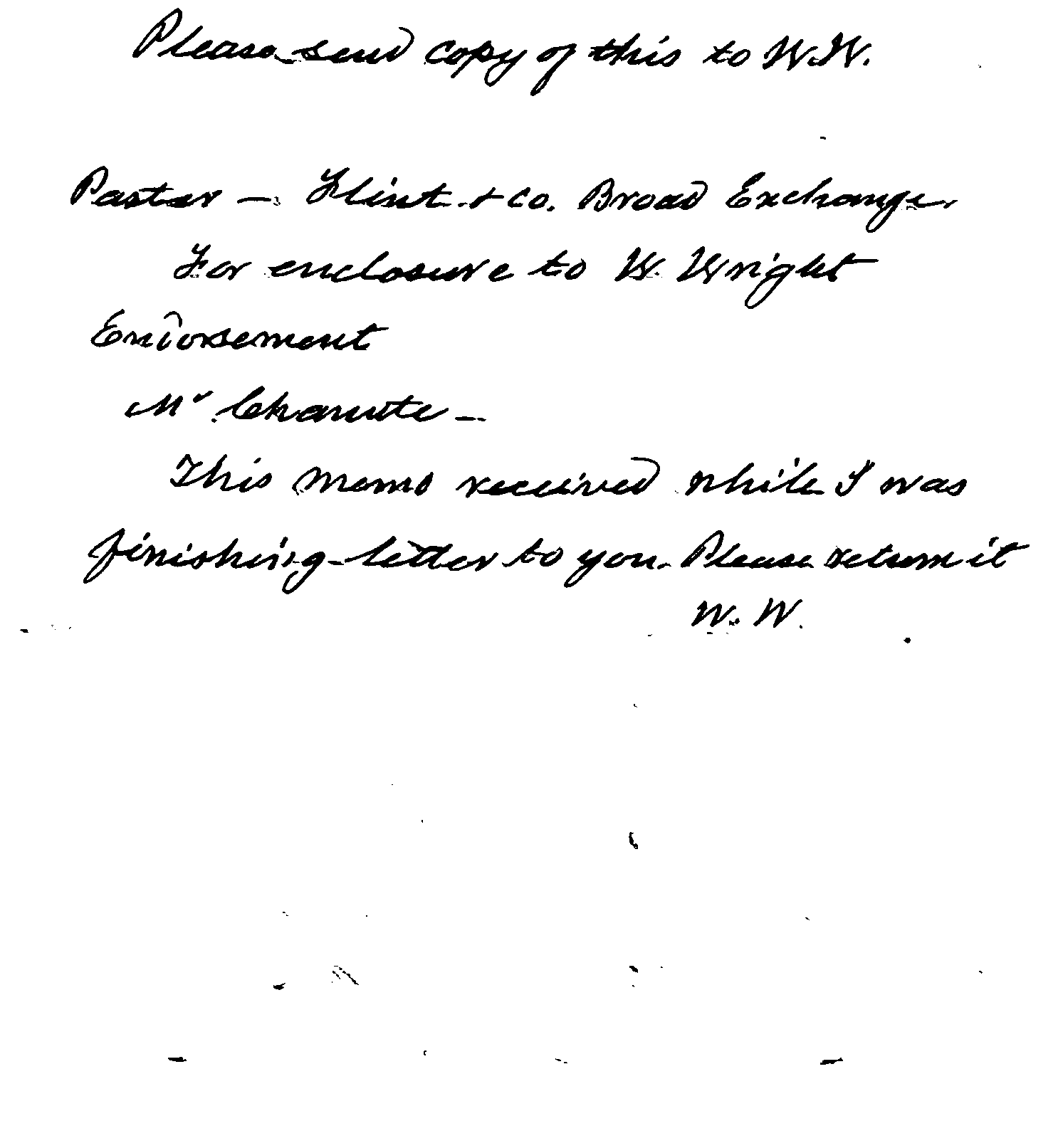}
		}
		\hskip 0.5\columnwidth
		\subfigure[]{
			\includegraphics[width=0.15\columnwidth]{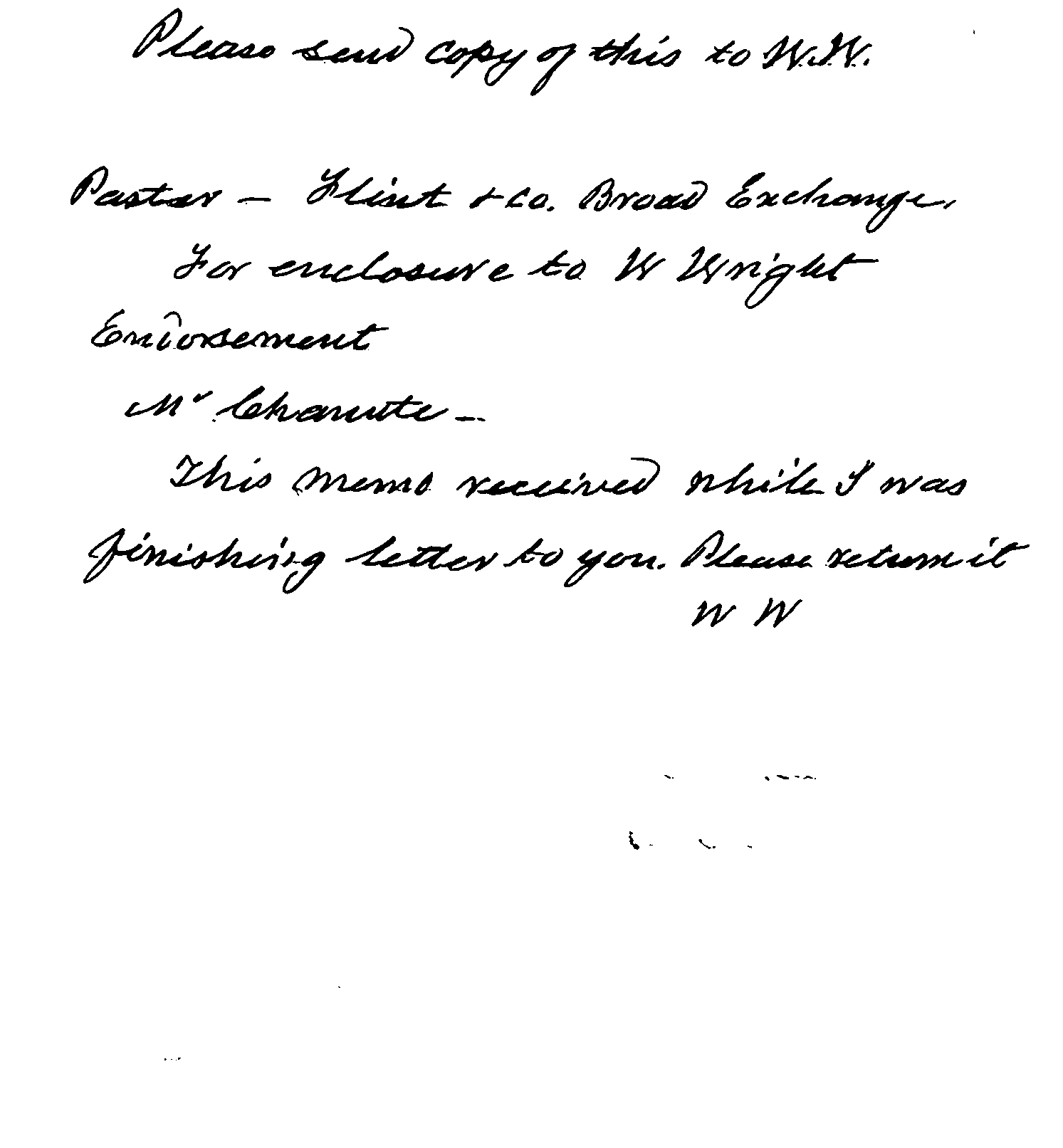}
		}
		\subfigure[]{
			\includegraphics[width=0.15\columnwidth]{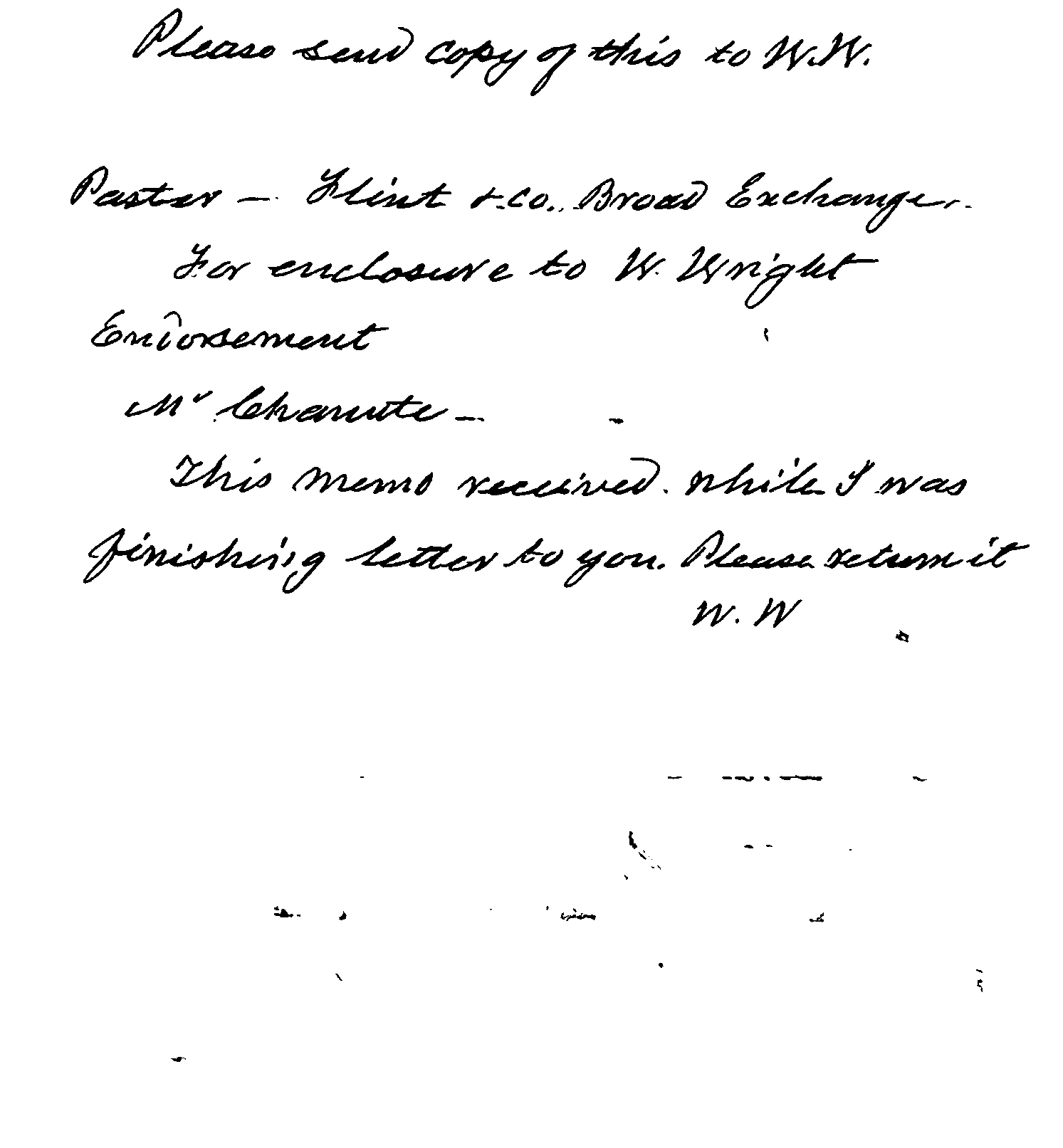}
		}
		\subfigure[]{
			\includegraphics[width=0.15\columnwidth]{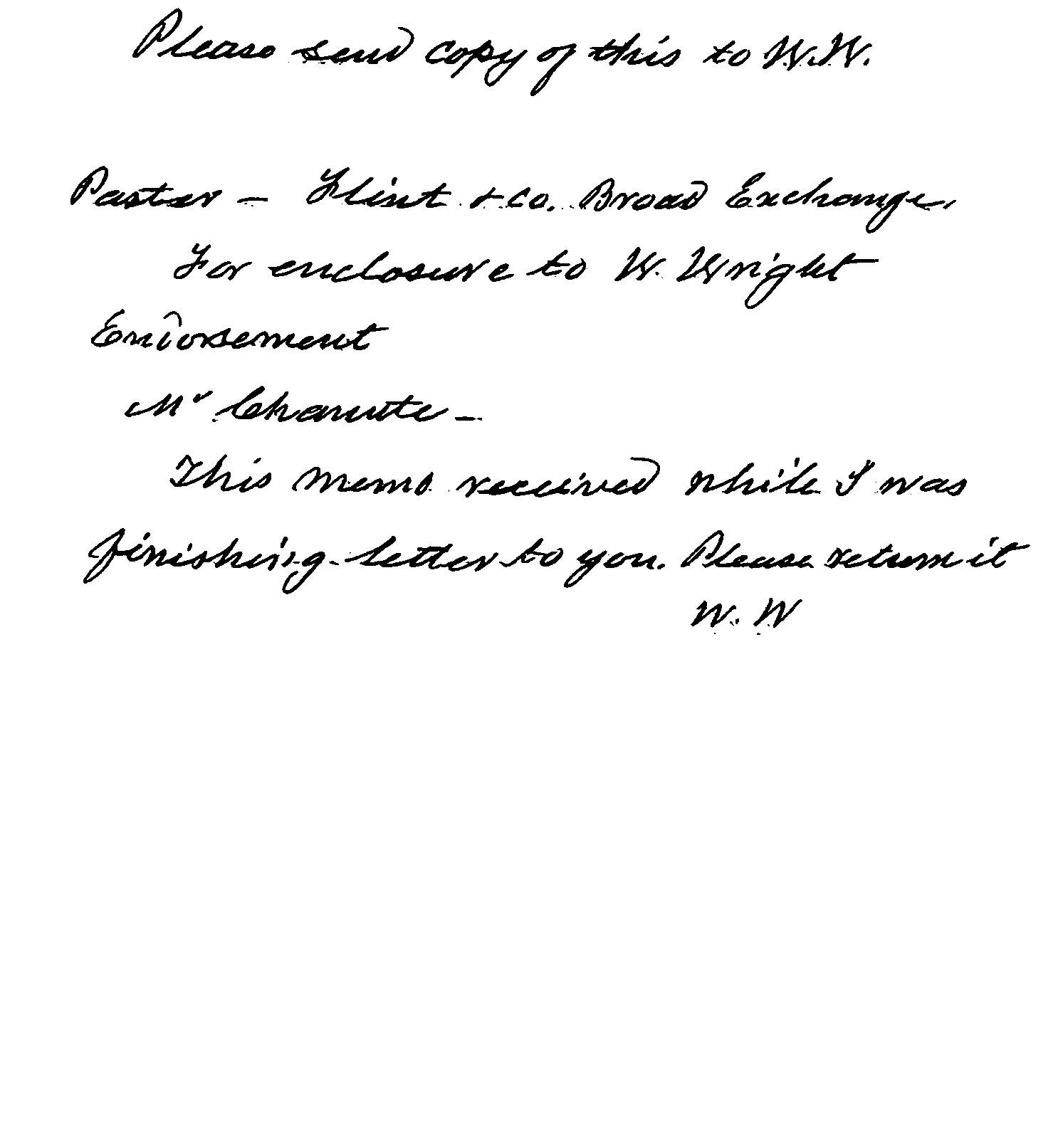}
		}
		\caption{Binarization results of the document image HW5 in DIBCO 2013. (a) original images, (b) the ground truth, (c) Otsu \cite{otsu1979threshold}, (d) Niblack \cite{niblack1986introduction}, (e) Sauvola \cite{sauvola2000adaptive}, (f) Vo \cite{vo2018binarization}, (g) He \cite{he2019deepotsu}, (h) Zhao \cite{zhao2019document}, (i) Ours.}
		\label{fig:DIBCO2013_5}
	\end{figure}
	\begin{figure}
		\centering
		\subfigure[]{
			\includegraphics[width=0.25\columnwidth]{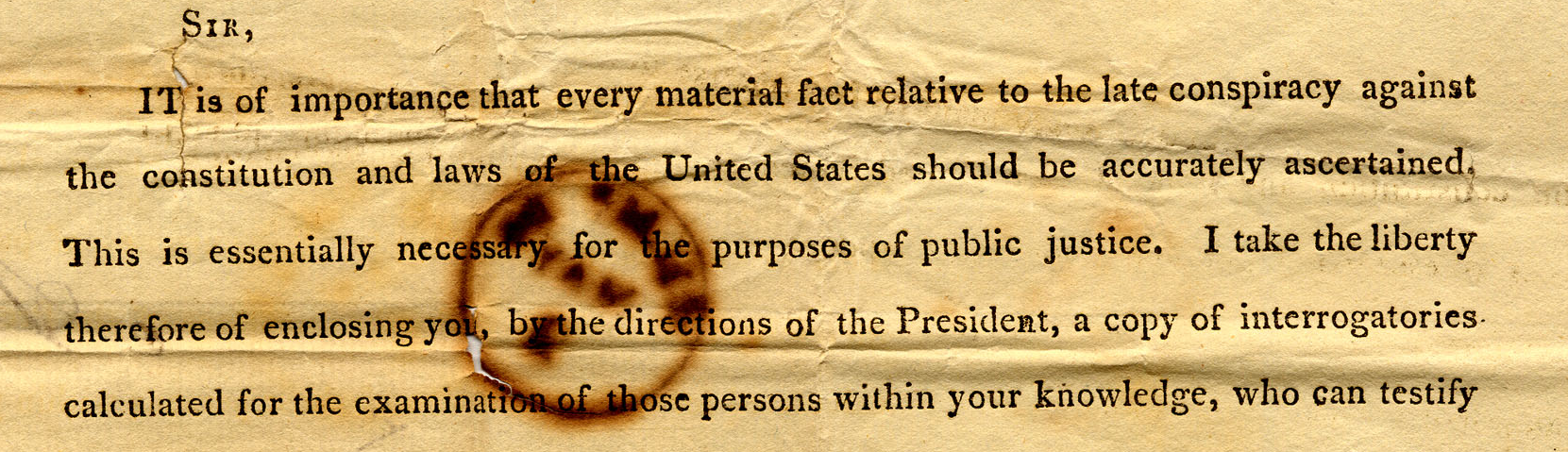}
		}
		\subfigure[]{
			\includegraphics[width=0.25\columnwidth]{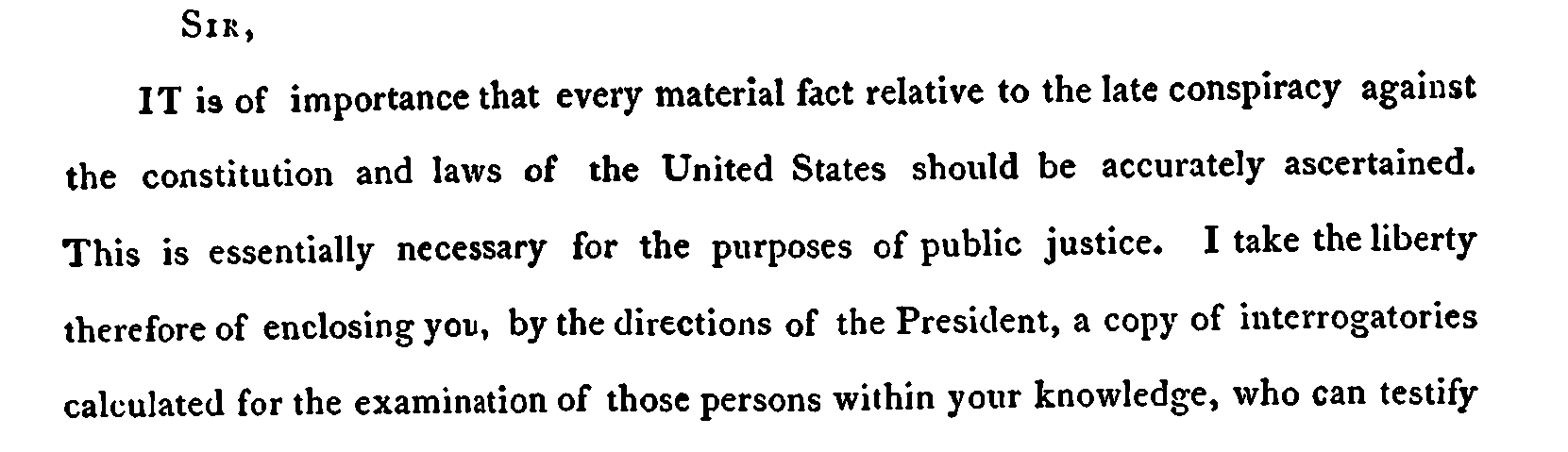}
		}
		\subfigure[]{
			\includegraphics[width=0.25\columnwidth]{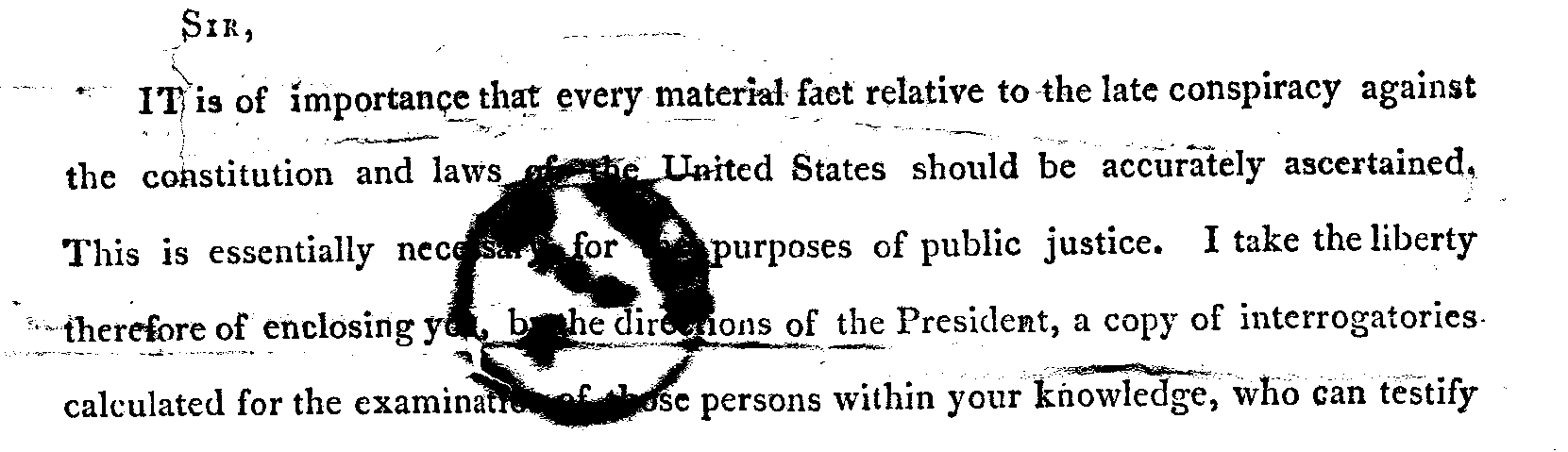}
		}
		\hfil
		\subfigure[]{
			\includegraphics[width=0.25\columnwidth]{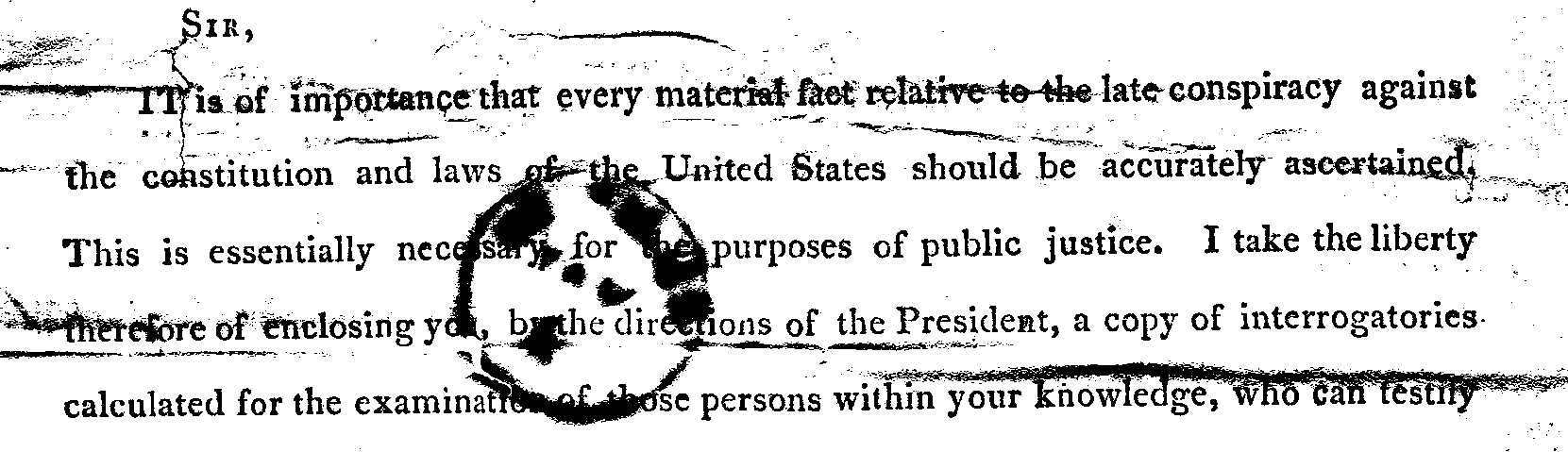}
		}
		\subfigure[]{
			\includegraphics[width=0.25\columnwidth]{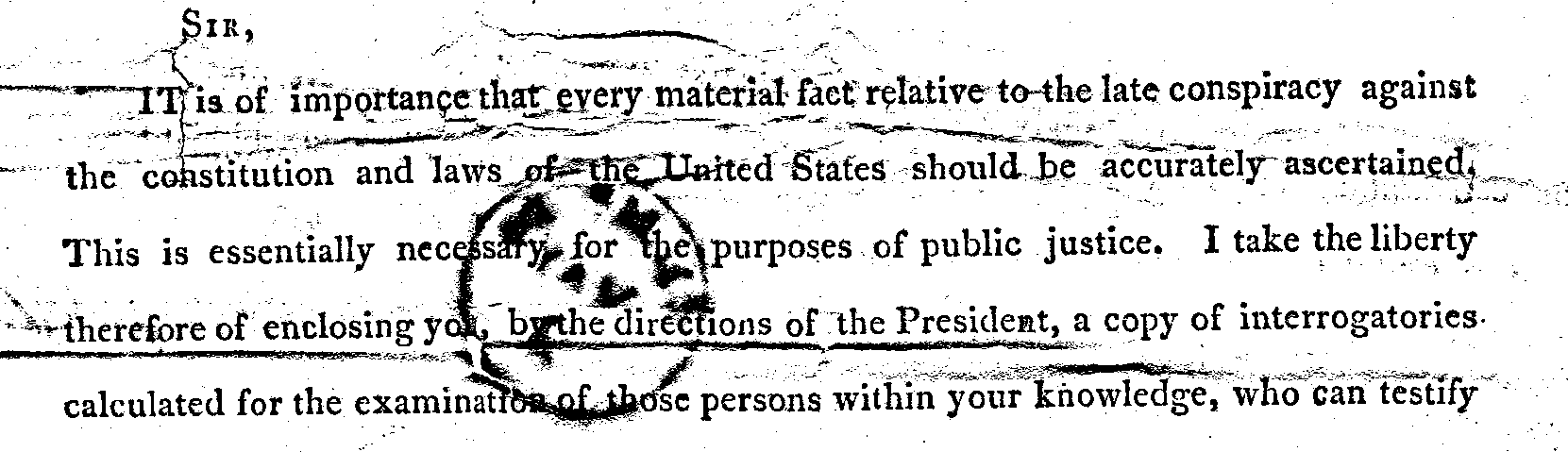}
		}
		\subfigure[]{
			\includegraphics[width=0.25\columnwidth]{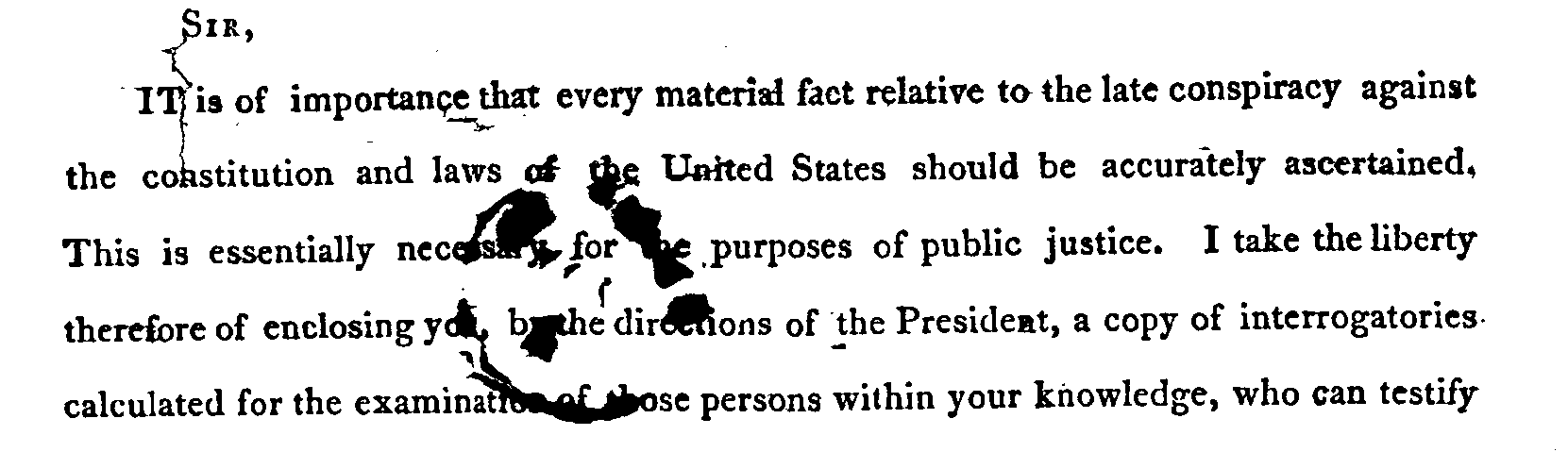}
		}
		\hfil
		\subfigure[]{
			\includegraphics[width=0.25\columnwidth]{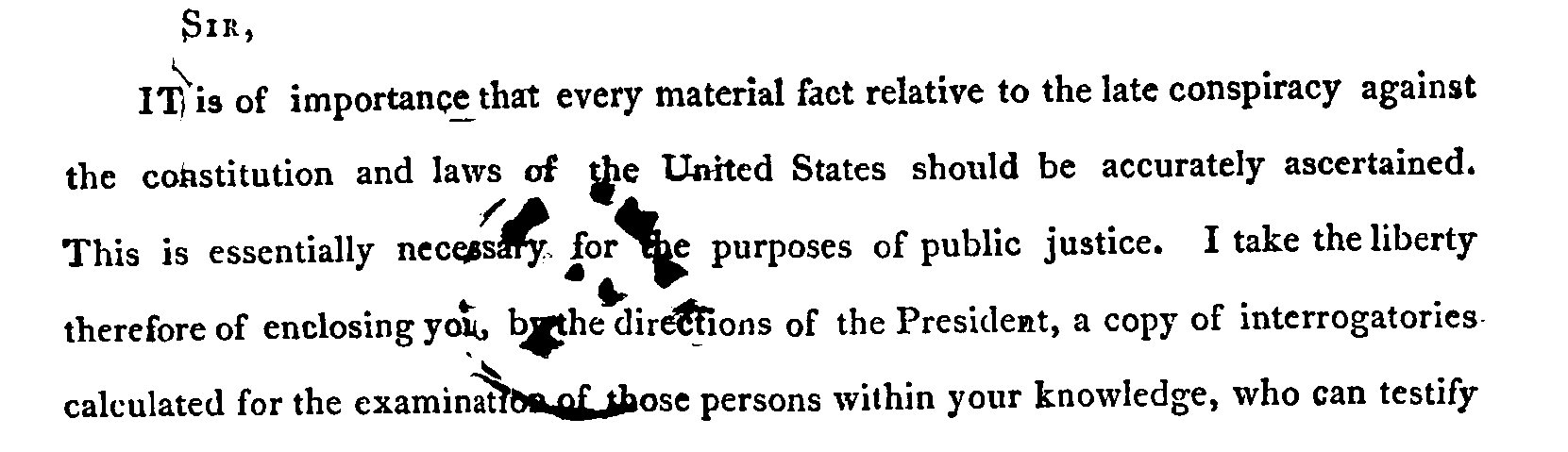}
		}
		\subfigure[]{
			\includegraphics[width=0.25\columnwidth]{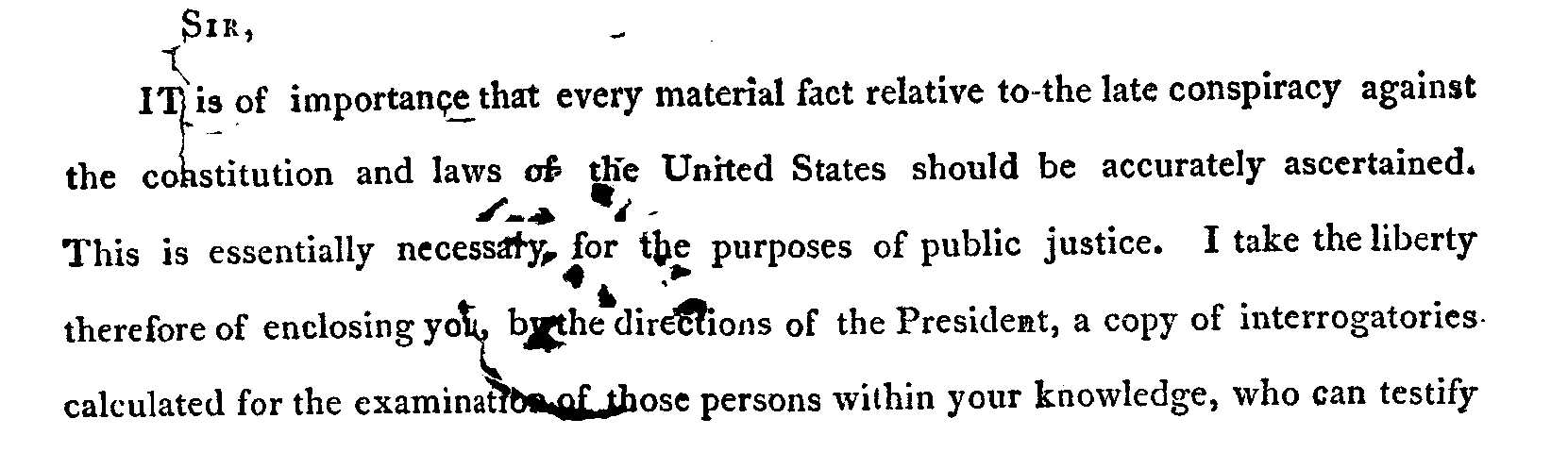}
		}
		\subfigure[]{
			\includegraphics[width=0.25\columnwidth]{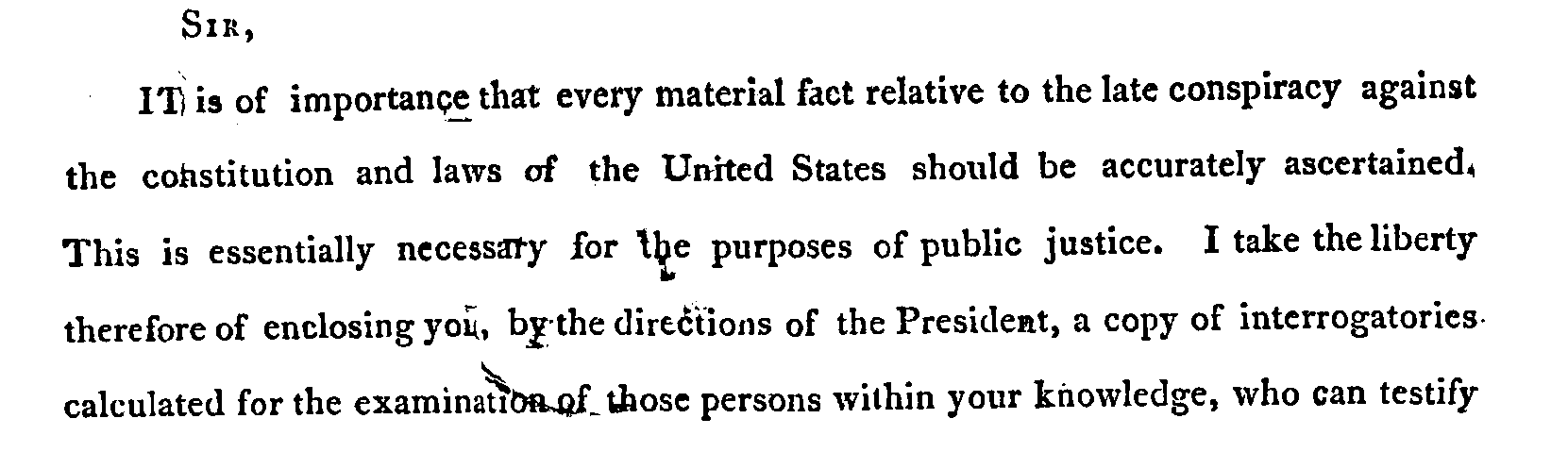}
		}
		\caption{Binarization results of the document image PR16 in DIBCO 2013. (a) original images, (b) the ground truth, (c) Otsu \cite{otsu1979threshold}, (d) Niblack \cite{niblack1986introduction}, (e) Sauvola \cite{sauvola2000adaptive}, (f) Vo \cite{vo2018binarization}, (g) He \cite{he2019deepotsu}, (h) Zhao \cite{zhao2019document}, (i) Ours.}
		\label{fig:DIBCO2013_16}
	\end{figure}

	Figure \ref{fig:DIBCO2011_1}, \ref{fig:DIBCO2013_5}, and \ref{fig:DIBCO2013_16} show the binary results produced by different methods on the two example images PR16 from DIBCO 2013. The deep-learning-based state-of-the-art methods can efficiently remove the shadow and text-like background noise better than the traditional binarization methods. The proposed method not only removes the shadow and noise better but also preserves the textual components.
		
	\subsubsection{Results on LRDE DBD}
	
	\begin{table}[!t]
		\caption{Evaluation of document image binarization on the LRDE DBD. }
		\label{tab:LRDEDBD}
		\centering
		\begin{tabular}{ccccc}
			\hline
			Methods & FM & p-FM & $PSNR$ & $DRD$\\
			\hline
			Otsu \cite{otsu1979threshold} & 91.12 & 97.33 & 19.70 & 2.59\\
			Niblack \cite{niblack1986introduction} & 84.33 & 87.00 & 17.66 & 10.72 \\
			Sauvola \cite{sauvola2000adaptive} & 85.34 & 96.23 & 17.72 & 6.22 \\
			Vo \cite{vo2018binarization} & 97.72 & \textbf{98.44} & 26.21 & 0.81 \\
			He \cite{he2019deepotsu} & 97.77 & 98.04 & 27.31 & 0.81 \\
			Zhao \cite{zhao2019document} & 97.97 & 98.42 & 27.39 & 0.75 \\
			Ours & \textbf{98.01} & 98.33 & \textbf{27.79} & \textbf{0.73}\\
			\hline
		\end{tabular}
	\end{table}
	Table \ref{tab:LRDEDBD} shows the mean values of the evaluation measures across the five-fold cross-validation procedure. The mean values are much higher than the values for the DIBCO datasets because the document images in LRDE DBD have color text and background without degradation. On LRDE DBD, the differences between the proposed method and the state-of-the-art methods are not large, but the proposed method achieved the best performance in terms of FM, PSNR, and DRD, whereas it ranked third in terms of p-FM. 
	
	\subsubsection{Results on Shipping Label Image Dataset}
	
	The shipping label image dataset contains images of shipping labels with the receiver's address area and the ground truth of the addresses and their text regions. The ground truth is manually masked and can be extracted for validation of the text recognition. In this subsection, we evaluate the proposed method and the state-of-the-art methods by the document image binarization metrics and the Levenshtein distance expressed in percent to measure the extent to which the improved binarization affects the quality of text recognition. For OCR of the text in the shipping addresses, we utilized Tesseract \cite{smith2007overview}, which recognizes characters based on long short-term memory (LSTM) \cite{hochreiter1997long}. This engine supports a very large database with 116 different languages. The shipping label image dataset was collected from eight countries and can be separated into two groups: one of labels written in the Latin alphabet and the other in Korean. The results of the OCR for the binarized address images were evaluated separately for the two language types.
	
	\begin{table}[!t]
		\caption{Evaluation of document image binarization on shipping label image dataset.}
		\label{tab:shippinglabel}
		\centering
		\begin{tabular}{ccccc}
			\hline
			Methods & FM & p-FM & $PSNR$ & $DRD$\\
			\hline
			Otsu \cite{otsu1979threshold} & 88.31 & 89.42 & 14.73 & 6.17\\
			Niblack \cite{niblack1986introduction} & 86.61 & 89.46 & 13.59 & 6.61 \\
			Sauvola \cite{sauvola2000adaptive} & 87.67 & 89.53 & 14.18 & 5.75 \\
			Vo \cite{vo2018binarization} & 91.20 & 92.92 & 16.14 & 2.20 \\
			He \cite{he2019deepotsu} & 91.09 & 92.26 & 16.03 & 2.33 \\
			Zhao \cite{zhao2019document} & 92.09 & 93.83 & 16.29 & 2.37 \\
			Ours & \textbf{94.65} & \textbf{95.94} & \textbf{18.02} & \textbf{1.57}\\
			\hline
		\end{tabular}
	\end{table}
	
	\begin{figure}
		\centering
		\subfigure[]{
			\includegraphics[width=0.3\columnwidth]{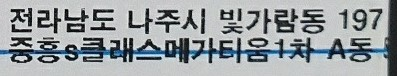}
		}
		\subfigure[]{
			\includegraphics[width=0.3\columnwidth]{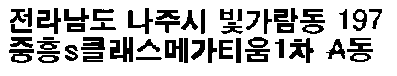}
		}
		\subfigure[]{
			\includegraphics[width=0.3\columnwidth]{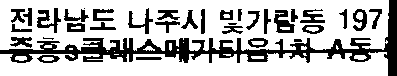}
		}
		\hfil
		\subfigure[]{
			\includegraphics[width=0.3\columnwidth]{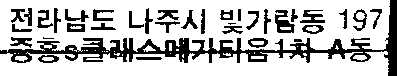}
		}
		\subfigure[]{
			\includegraphics[width=0.3\columnwidth]{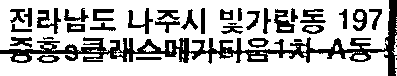}
		}
		\subfigure[]{
			\includegraphics[width=0.3\columnwidth]{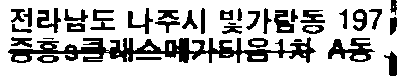}
		}
		\hfil
		\subfigure[]{
			\includegraphics[width=0.3\columnwidth]{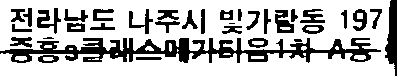}
		}
		\subfigure[]{
			\includegraphics[width=0.3\columnwidth]{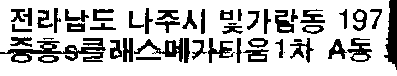}
		}
		\subfigure[]{
			\includegraphics[width=0.3\columnwidth]{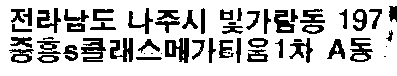}
		}
		\caption{Binarization results of the address image Kor230 from the shipping label image dataset. (a) original images, (b) the ground truth, (c) Otsu \cite{otsu1979threshold}, (d) Niblack \cite{niblack1986introduction}, (e) Sauvola \cite{sauvola2000adaptive}, (f) Vo \cite{vo2018binarization}, (g) He \cite{he2019deepotsu}, (h) Zhao \cite{zhao2019document}, (i) Ours.}
		\label{fig:ShippingLabelKor230}
	\end{figure}
	\begin{figure}
		\centering
		\subfigure[]{
			\includegraphics[width=0.3\columnwidth]{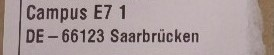}
		}
		\subfigure[]{
			\includegraphics[width=0.3\columnwidth]{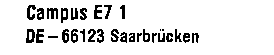}
		}
		\subfigure[]{
			\includegraphics[width=0.3\columnwidth]{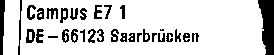}
		}
		\hfil
		\subfigure[]{
			\includegraphics[width=0.3\columnwidth]{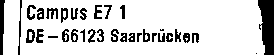}
		}
		\subfigure[]{
			\includegraphics[width=0.3\columnwidth]{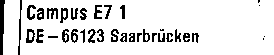}
		}
		\subfigure[]{
			\includegraphics[width=0.3\columnwidth]{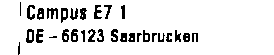}
		}
		\hfil
		\subfigure[]{
			\includegraphics[width=0.3\columnwidth]{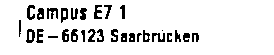}
		}
		\subfigure[]{
			\includegraphics[width=0.3\columnwidth]{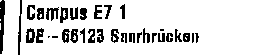}
		}
		\subfigure[]{
			\includegraphics[width=0.3\columnwidth]{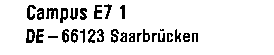}
		}
		\caption{Binarization results of the address image Ger142 from the shipping label image dataset. (a) original images, (b) the ground truth, (c) Otsu, (d) Niblack, (e) Sauvola, (f) Vo, (g) He, (h) Zhao, (i) Ours.}
		\label{fig:ShippingLabelGermany142}
	\end{figure}
	\begin{figure}
		\centering
		\subfigure[]{
			\includegraphics[width=0.3\columnwidth]{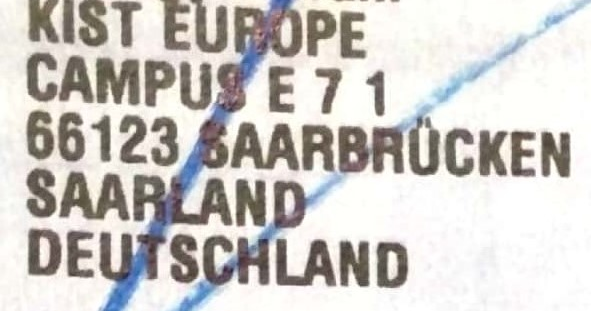}
		}
		\subfigure[]{
			\includegraphics[width=0.3\columnwidth]{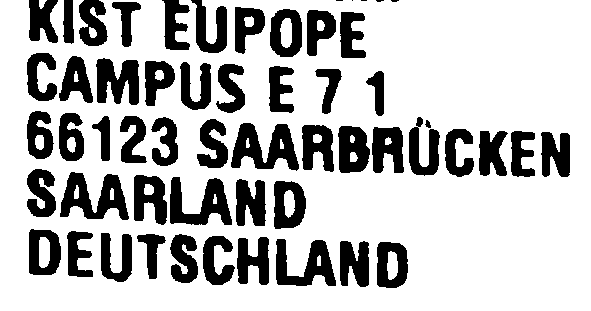}
		}
		\subfigure[]{
			\includegraphics[width=0.3\columnwidth]{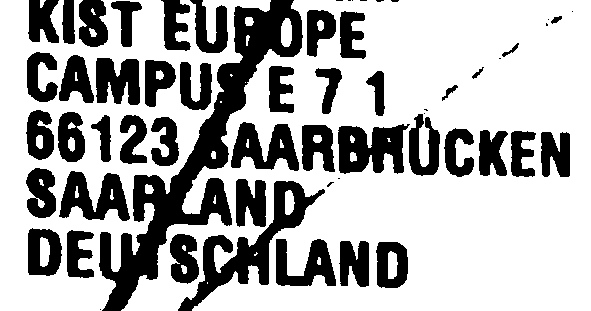}
		}
		\hfil
		\subfigure[]{
			\includegraphics[width=0.3\columnwidth]{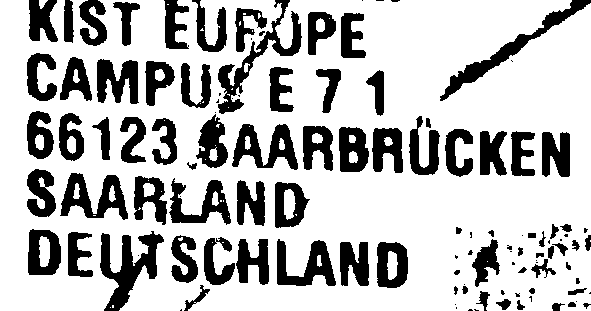}
		}
		\subfigure[]{
			\includegraphics[width=0.3\columnwidth]{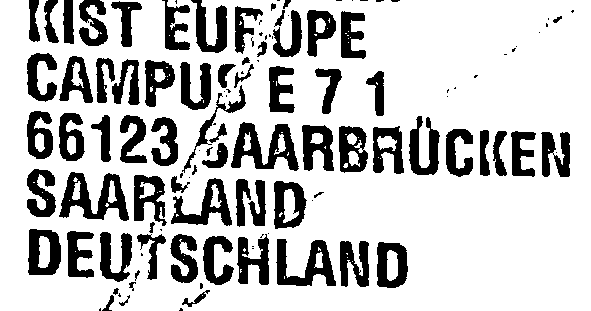}
		}
		\subfigure[]{
			\includegraphics[width=0.3\columnwidth]{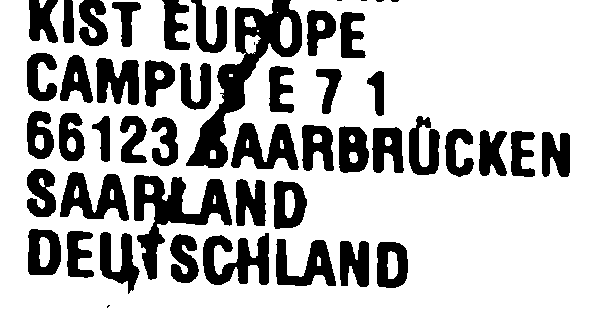}
		}
		\hfil
		\subfigure[]{
			\includegraphics[width=0.3\columnwidth]{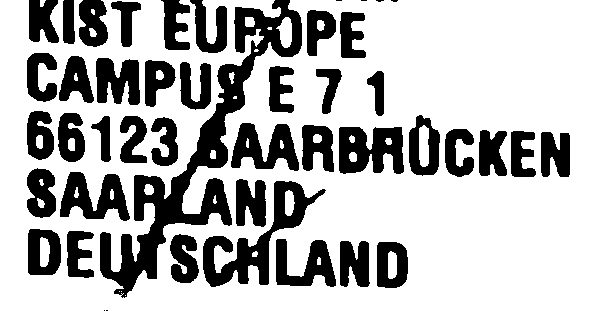}
		}
		\subfigure[]{
			\includegraphics[width=0.3\columnwidth]{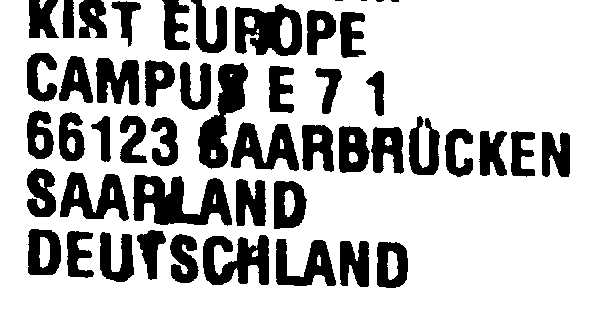}
		}
		\subfigure[]{
			\includegraphics[width=0.3\columnwidth]{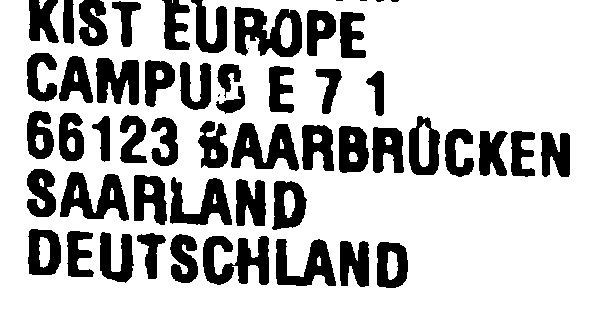}
		}
		\caption{Binarization results of the address image Ger177 from the shipping label image dataset. (a) original images, (b) the ground truth, (c) Otsu \cite{otsu1979threshold}, (d) Niblack \cite{niblack1986introduction}, (e) Sauvola \cite{sauvola2000adaptive}, (f) Vo \cite{vo2018binarization}, (g) He \cite{he2019deepotsu}, (h) Zhao \cite{zhao2019document}, (i) Ours.}
		\label{fig:ShippingLabelGermany177}
	\end{figure}
	\begin{figure}
		\centering
		\subfigure[]{
			\includegraphics[width=0.3\columnwidth]{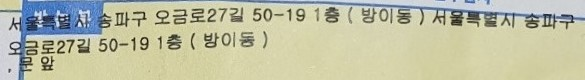}
		}
		\subfigure[]{
			\includegraphics[width=0.3\columnwidth]{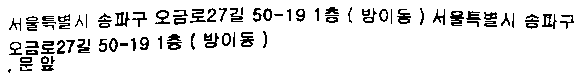}
		}
		\subfigure[]{
			\includegraphics[width=0.3\columnwidth]{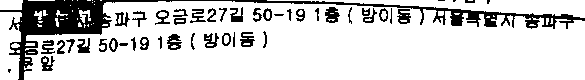}
		}
		\hfil
		\subfigure[]{
			\includegraphics[width=0.3\columnwidth]{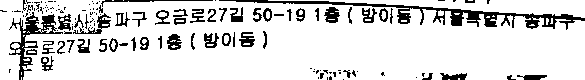}
		}
		\subfigure[]{
			\includegraphics[width=0.3\columnwidth]{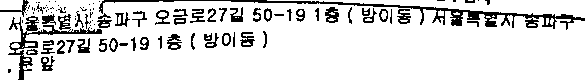}
		}
		\subfigure[]{
			\includegraphics[width=0.3\columnwidth]{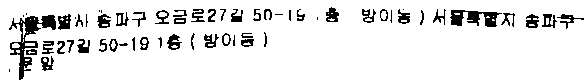}
		}
		\hfil
		\subfigure[]{
			\includegraphics[width=0.3\columnwidth]{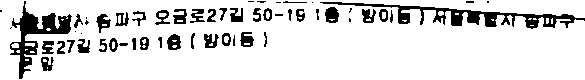}
		}
		\subfigure[]{
			\includegraphics[width=0.3\columnwidth]{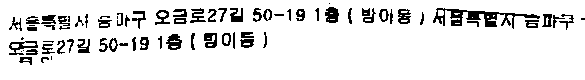}
		}
		\subfigure[]{
			\includegraphics[width=0.3\columnwidth]{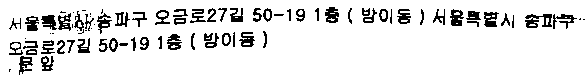}
		}
		\caption{Binarization results of the address image Kor19 from the shipping label image dataset. (a) original images, (b) the ground truth, (c) Otsu \cite{otsu1979threshold}, (d) Niblack \cite{niblack1986introduction}, (e) Sauvola \cite{sauvola2000adaptive}, (f) Vo \cite{vo2018binarization}, (g) He \cite{he2019deepotsu}, (h) Zhao \cite{zhao2019document}, (i) Ours.}
		\label{fig:ShippingLabelKor19}
	\end{figure}	
	\begin{figure}
		\centering
		\subfigure[]{
			\includegraphics[width=0.3\columnwidth]{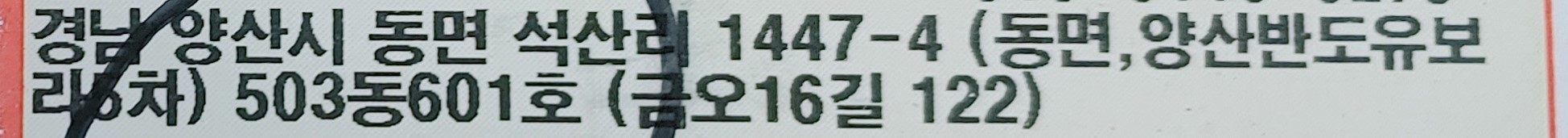}
		}
		\subfigure[]{
			\includegraphics[width=0.3\columnwidth]{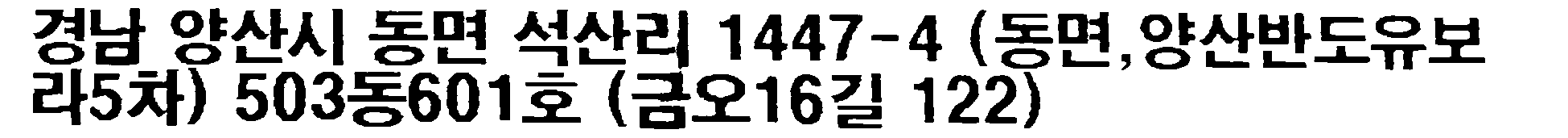}
		}
		\subfigure[]{
			\includegraphics[width=0.3\columnwidth]{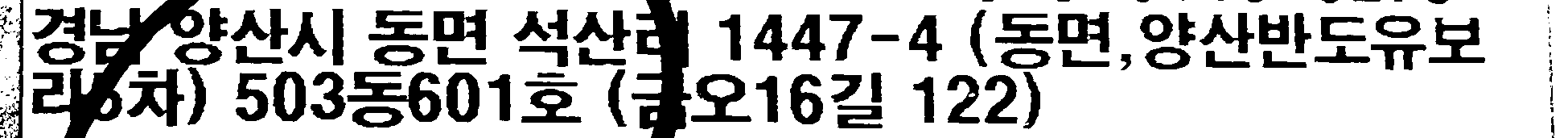}
		}
		\hfil
		\subfigure[]{
			\includegraphics[width=0.3\columnwidth]{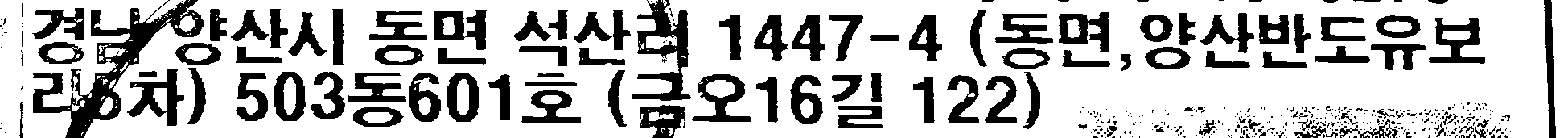}
		}
		\subfigure[]{
			\includegraphics[width=0.3\columnwidth]{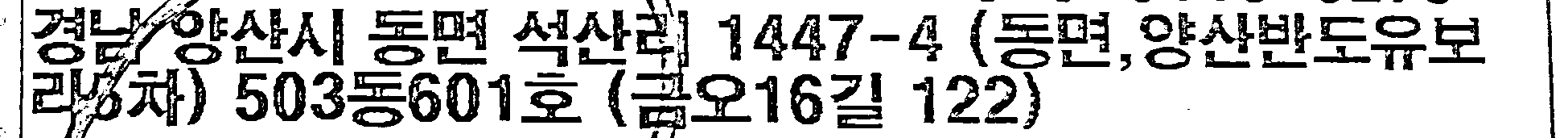}
		}
		\subfigure[]{
			\includegraphics[width=0.3\columnwidth]{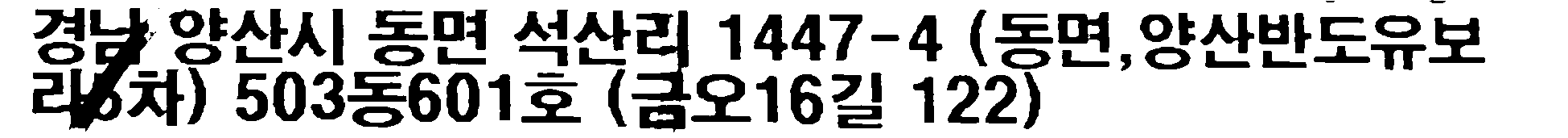}
		}
		\hfil
		\subfigure[]{
			\includegraphics[width=0.3\columnwidth]{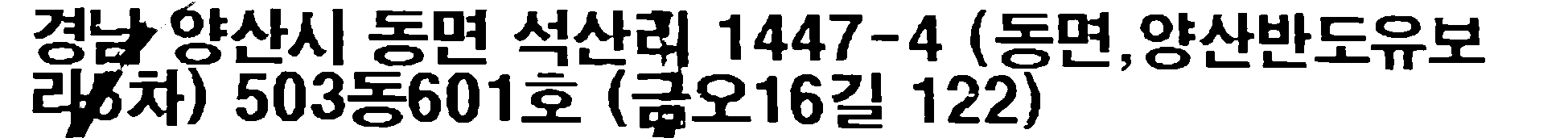}
		}
		\subfigure[]{
			\includegraphics[width=0.3\columnwidth]{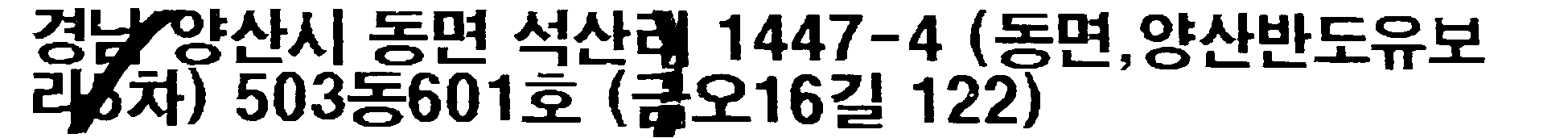}
		}
		\subfigure[]{
			\includegraphics[width=0.3\columnwidth]{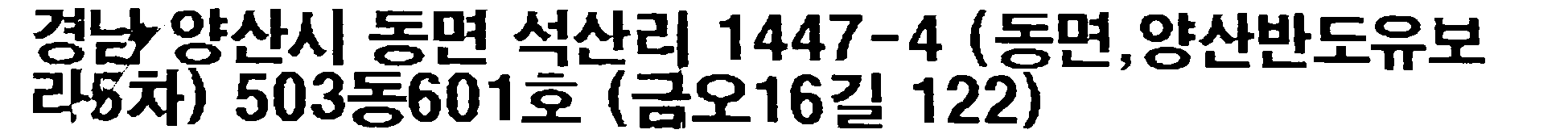}
		}
		\caption{Binarization results of the address image Kor265 from the shipping label image dataset. (a) original images, (b) the ground truth, (c) Otsu \cite{otsu1979threshold}, (d) Niblack \cite{niblack1986introduction}, (e) Sauvola \cite{sauvola2000adaptive}, (f) Vo \cite{vo2018binarization}, (g) He \cite{he2019deepotsu}, (h) Zhao \cite{zhao2019document}, (i) Ours.}
		\label{fig:ShippingLabelKor265}
	\end{figure}

	\begin{table}[!t]
		\caption{OCR accuracy comparison in Levenshtein distance in percent on the shipping label image dataset.}
		\label{table:recognitionresults}
		\centering
		\begin{tabular}{cccc}
			\hline
			Methods & Total & Korean & Alphabet \\
			\hline
			Input image & 77.20 & 73.86 & 94.47 \\
			Ground Truth & 87.62 & 85.88 & 96.66 \\
			Otsu \cite{otsu1979threshold} & 74.45 & 70.72 & 93.79\\
			Niblack \cite{niblack1986introduction} & 69.00 & 66.31 & 82.94 \\
			Sauvola \cite{sauvola2000adaptive} & 72.84 & 68.81 & 93.73 \\
			Vo \cite{vo2018binarization} & 77.14 & 74.69 & 89.86 \\
			He \cite{he2019deepotsu} & 75.15 & 72.45 & 89.13 \\
			Zhao \cite{zhao2019document} & 77.33 & 74.56 & 91.69 \\
			Ours & \textbf{83.40} & \textbf{81.15} & \textbf{95.09}\\
			\hline
		\end{tabular} 
	\end{table}
	
	Table \ref{tab:shippinglabel} shows the mean values of the evaluation measures across the five-fold cross-validation procedure. 
	Even though the shipping label dataset images contain colorful backgrounds and contaminants with colors unlike the other datasets, the proposed method outperformed the other methods in terms of all of the metrics.
	Figure \ref{fig:ShippingLabelKor230}, \ref{fig:ShippingLabelGermany142}, \ref{fig:ShippingLabelGermany177}, \ref{fig:ShippingLabelKor19}, and \ref{fig:ShippingLabelKor265} show results of the image binarization and text region extraction; it can be seen that the deep-learning-based methods can extract the text regions from the background regions, in contrast to the traditional binarization methods. The proposed method can remove the colorful backgrounds and contaminants in the shipping label images, whereas the state-of-the-art methods handle the color information improperly. Table \ref{table:recognitionresults} shows the average OCR accuracy results using Levenshtein distance expressed in percent. The proposed method improved the accuracy of the input images by 6.20 percentage points. Moreover, the proposed method provided the best performance on both the Korean and Latin alphabet datasets.

	\begin{table}[!t]
		\caption{Average runtime of different methods on the shipping label image dataset in milliseconds}
		\label{tab:runtime}
		\centering
		\begin{tabular}{cc}
			\hline
			Method & Runtime\\
			\hline
			Otsu \cite{otsu1979threshold} 		& 1.5 \\
			Niblack \cite{niblack1986introduction} 	& 350.2 \\
			Sauvola \cite{sauvola2000adaptive}		& 15.1 \\
			Vo \cite{vo2018binarization}		& 139.0 \\
			He \cite{he2019deepotsu}	& 856.2 \\
			Zhao \cite{zhao2019document}	& 131.8 \\
			Ours & 282.8\\
			\hline
		\end{tabular}
	\end{table}
	
	To measure the efficiency, the run time on the shipping label image dataset was computed. Table \ref{tab:runtime} shows the average run time of the different methods on the shipping label image dataset. The average run time of the deep-learning-based methods was much greater than that of the traditional image processing methods. Of the deep-learning-based methods, the proposed method took more time than Vo et al.'s and Zhao et al.'s methods but was significantly faster than He and Schomaker's method.
	
	\subsection{Model Ablation Study}
	Although the comparison experiments showed that the proposed method provided better performance than state-of-the-art methods with the converted grayscale image or the original three-channel images as the input to the neural networks, we would like to better understand the benefits brought by the proposed method. Here, we performed our ablation study to verify the advantage of individual modules in the proposed model. 
	
	To verify the benefit of the proposed two-stage framework and the fusion of the multi-scale information by using the resized original image and the patches from the first stage output image, we derive two variants of the proposed method, named ‘Step1-Only’ and ‘NoOriginal’. In particular, ‘Step1-Only’ refers to the non-adapted version of the proposed second stage, and ‘NoOriginal’ is implemented with the resized first stage output image instead of the resized original image for the global binarization.
	
	Table \ref{tab:ablationstudy} shows the comparison between the proposed method and its two variants on the four DIBCO datasets. The ‘Step1-Only’ provided the worst performance and the proposed method outperformed the ‘Step1-Only’ and ‘NoOriginal’ in terms of all four of the measurements, which signifies the effectiveness of the two-stage design and the usage of the resized original image for the global binarization.
		
	\begin{table}[!t]
		\caption{Ablation study of the proposed method.}
		\label{tab:ablationstudy}
		\centering
		\subfigure[DIBCO 2011]{
			\resizebox{0.45\columnwidth}{!}{
				\begin{tabular}{ccccc}
					\hline
					Methods & FM & p-FM & $PSNR$ & $DRD$\\
					\hline
					Step1-Only	& 93.09 & 94.61 & 19.83 & 2.51 \\
					NoOriginal  & 93.42 & 95.69 & 20.07 & 2.20 \\
					NoGAN		& 93.43 & 95.90 & 20.13 & 1.98 \\
					Proposed 	& \textbf{93.81} & \textbf{96.04} & \textbf{20.26} & \textbf{1.96}\\
					\hline
			\end{tabular}}
		}
		\subfigure[DIBCO 2013]{
			\resizebox{0.45\columnwidth}{!}{
				\begin{tabular}{ccccc}
					\hline
					Methods & FM & p-FM & $PSNR$ & $DRD$\\
					\hline
					Step1-Only	& 93.67 & 94.54 & 21.07 & 2.78 \\
					NoOriginal  & 94.16 & 95.50 & 21.50 & 2.42 \\
					NoGAN		& 94.18 & 94.91 & 21.44 & 2.51 \\
					Proposed 	& \textbf{94.82} & \textbf{96.14} & \textbf{21.89} & \textbf{1.88}\\
					\hline
			\end{tabular}}
		}
		\hfil
		\subfigure[H-DIBCO 2014]{
			\resizebox{0.45\columnwidth}{!}{
				\begin{tabular}{ccccc}
					\hline
					Methods & FM & p-FM & $PSNR$ & $DRD$\\
					\hline
					Step1-Only	& 96.48 & 97.45 & 21.98 & 1.05 \\
					NoOriginal  & 96.51 & 98.02 & 22.12 & 0.99 \\
					NoGAN		& 96.19 & 97.69 & 21.96 & 1.09 \\
					Proposed 	& \textbf{96.56} & \textbf{98.07} & \textbf{22.18} & \textbf{0.96}\\
					\hline
			\end{tabular}}
		}
		\subfigure[H-DIBCO 2016]{
			\resizebox{0.45\columnwidth}{!}{
				\begin{tabular}{ccccc}
					\hline
					Methods & FM & p-FM & $PSNR$ & $DRD$\\
					\hline
					Step1-Only	& 91.46 & 94.35 & 19.35 & 3.48 \\
					NoOriginal  & 91.75 & 95.59 & 19.63 & 3.10 \\
					NoGAN		& 91.79 & 95.55 & 19.74 & 2.99 \\
					Proposed 	& \textbf{92.08} & \textbf{95.89} & \textbf{19.85} & \textbf{2.87}\\
					\hline
			\end{tabular}}
		}
	\end{table}
	
	\begin{figure}[!t]
		\centering
		\subfigure[]{
			\includegraphics[width=0.17\columnwidth]{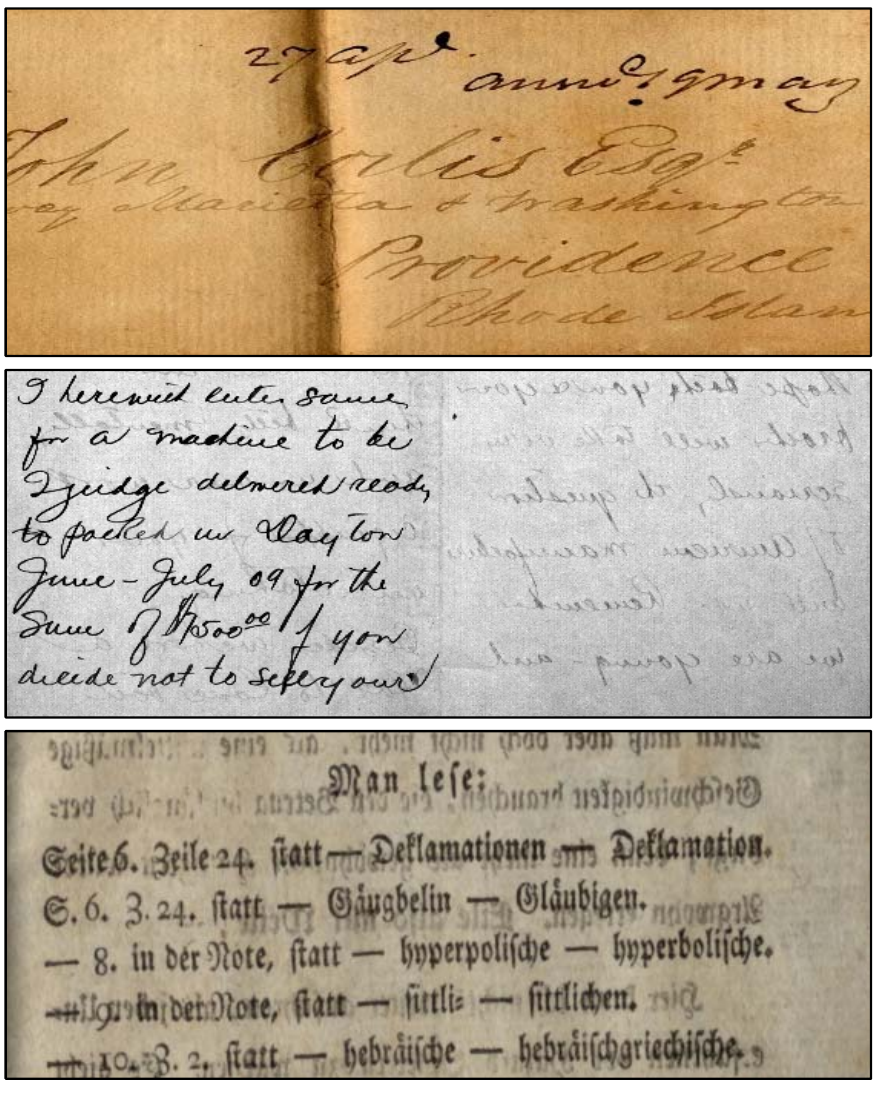}
		}
		\subfigure[]{
			\includegraphics[width=0.17\columnwidth]{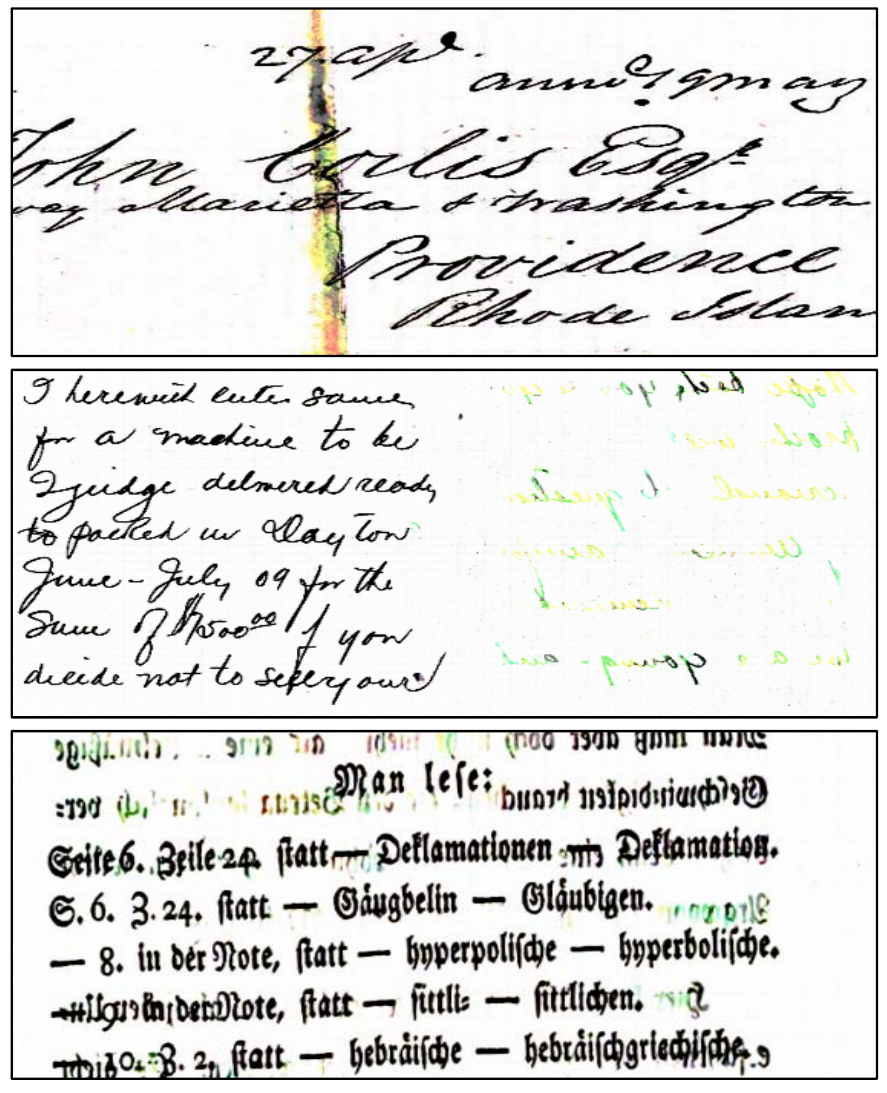}
		}
		\subfigure[]{
			\includegraphics[width=0.17\columnwidth]{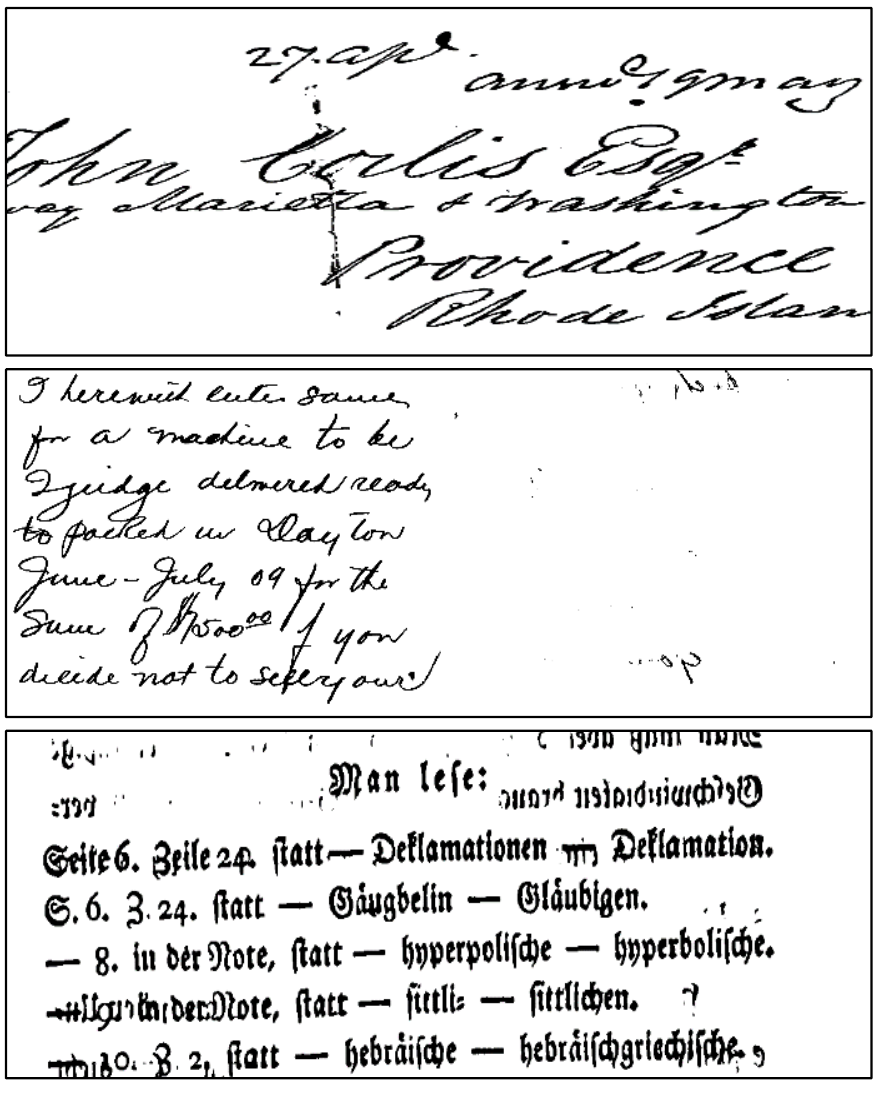}
		}
		\subfigure[]{
			\includegraphics[width=0.17\columnwidth]{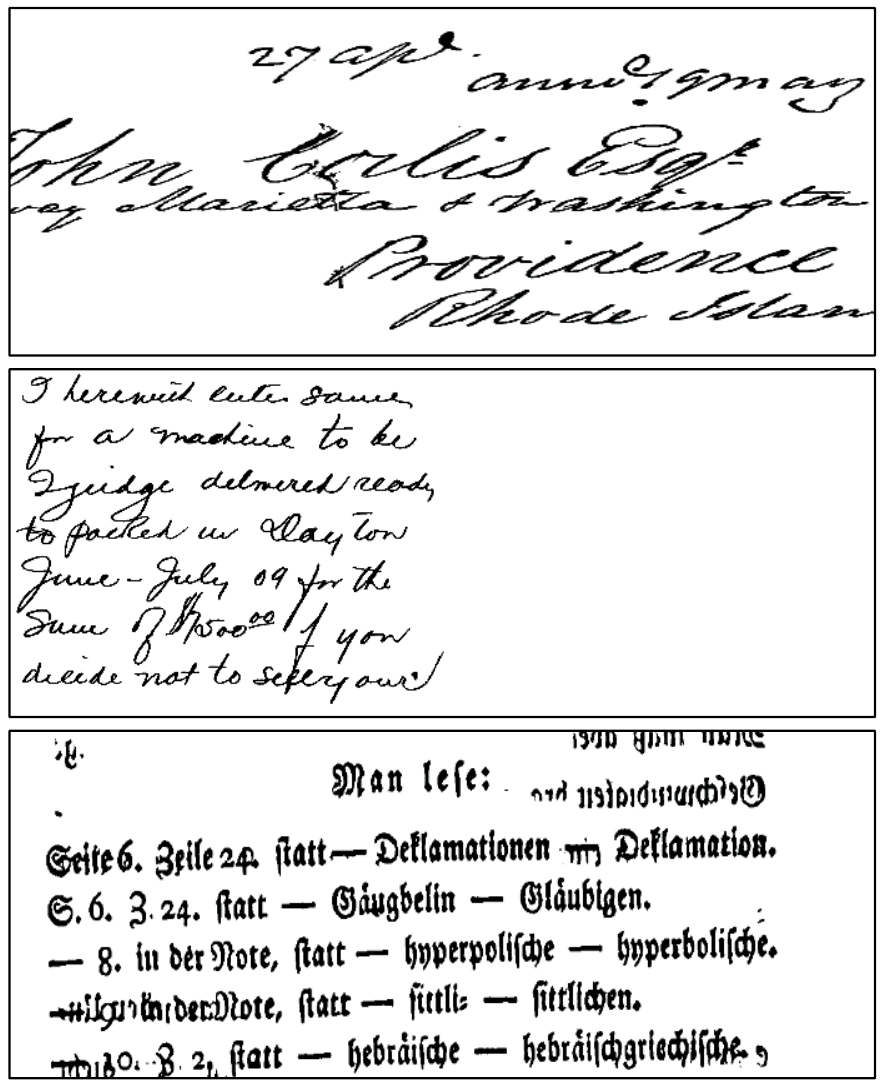}
		}
		\subfigure[]{
			\includegraphics[width=0.17\columnwidth]{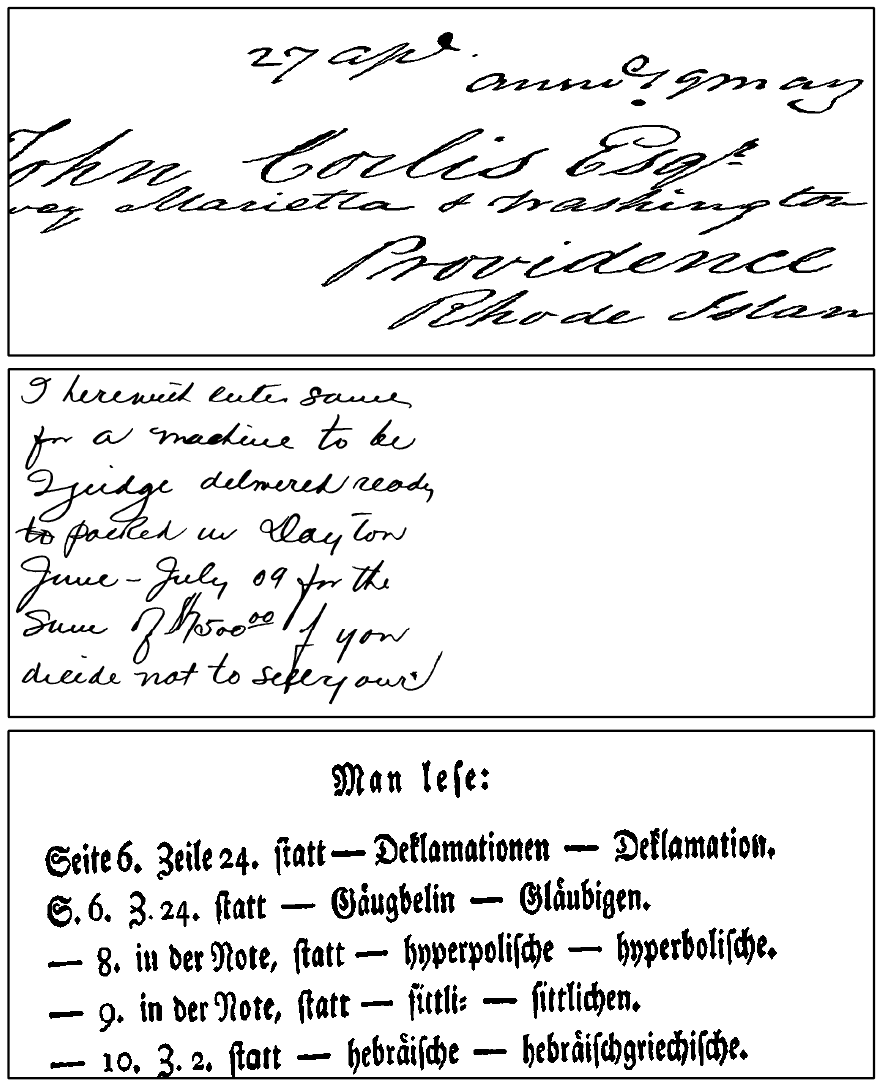}
		}
		\caption{Example images of the enhancement and binarization results of document images from DIBCO 2011 and DIBCO 2013. (a) original images, (b) the enhanced images using proposed image enhancement method (Stage 1), (c) the binarization images using proposed local prediction method without global prediction method (Stage 2), (d) the final binarization results using the proposed method combining the local and global features (Stage 2), (e) ground truth images. }
		\label{fig:DIBCOSteps}
	\end{figure}
	
	Figure \ref{fig:DIBCOSteps} shows the enhancement and binarization images obtained using the proposed method step by step on DIBCO 2011 and DIBCO 2013. The enhancement results from the proposed color-independent adversarial networks (Figure \ref{fig:DIBCOSteps}(b)) show that the background color has been removed and the color of the textual regions is highlighted. Comparing the binarization images produced by the proposed multi-scale binarization method (Figure \ref{fig:DIBCOSteps} parts (c) and (d)), we can see that the noise of the pure background patches have been effectively removed in Figure \ref{fig:DIBCOSteps}(d).  The results are close to the ground truth images (Figure \ref{fig:DIBCOSteps}(e)).

	\section{Conclusion} \label{sec:conclusion}
	
	In this paper, we have proposed a color document image enhancement and binarization method using multi-scale adversarial neural networks. The proposed method focuses on the multi-color degradation problem with its design consisting of four color-independent networks and combines the local and global binary transformation networks. We evaluated the proposed method on widely used DIBCO datasets, LRDE DBD, and a shipping label image dataset. The experimental results demonstrate that the proposed method outperforms traditional and state-of-the-art methods.
	
	In future work, we intend to integrate document image binarization with text recognition for practical applications. For instance, we can deploy the proposed method for automated shipping address recognition and validation based on the enhanced image binarization result, which will be applicable to the packaging machine and logistics industry. The shipping label image dataset we evaluated in this paper will be published on our website for use in further research by the computer vision community.

	\section*{Acknowledgments}
	This work was supported by the KIST Europe Institutional Program (Project No. 12120).

	\bibliography{mybibfile}	
\end{document}